\documentclass{article}

\usepackage{microtype}
\usepackage{graphicx}
\usepackage{subcaption}
\usepackage{booktabs} 
\usepackage{multirow} 

\usepackage{hyperref}

\usepackage[preprint]{icml2026}

\usepackage{amsmath}
\usepackage{amssymb}
\usepackage{mathtools}
\usepackage{amsthm}
\usepackage{cuted}

\usepackage{placeins} 

\usepackage{float}

\usepackage[capitalize,noabbrev]{cleveref}

\begin{document}
\theoremstyle{plain}
\newtheorem{theorem}{Theorem}[section]
\newtheorem{proposition}[theorem]{Proposition}
\newtheorem{lemma}[theorem]{Lemma}
\newtheorem{corollary}[theorem]{Corollary}
\theoremstyle{definition}
\newtheorem{definition}[theorem]{Definition}
\newtheorem{assumption}[theorem]{Assumption}
\theoremstyle{remark}
\newtheorem{remark}[theorem]{Remark}


\icmltitlerunning{Rethinking Time Series Domain Generalization via Structure-Stratified Calibration}

\twocolumn[
  \icmltitle{Rethinking Time Series Domain Generalization via Structure-Stratified Calibration}



  \icmlsetsymbol{equal}{*}

  \begin{icmlauthorlist}
    \icmlauthor{Jinyang Li}{equal,sch}
    \icmlauthor{Shuhao Mei}{equal,sch}
    \icmlauthor{Xiaoyu Xiao}{sch}
    \icmlauthor{Shuhang Li}{sch}
    \icmlauthor{Ruoxi Yun}{sch}
    \icmlauthor{Jinbo Sun}{sch}
  \end{icmlauthorlist}

  \icmlaffiliation{sch}{Guangzhou Institute of Technology, Xidian University, Xi’an, China}

  \icmlcorrespondingauthor{Jinbo Sun}{sunjb@xidian.edu.cn}

  \icmlkeywords{Time series, Cross-dataset generalization, Domain generalization, Latent dynamical systems, Spectral alignment, Structural heterogeneity}

  \vskip 0.3in
]



\printAffiliationsAndNotice{\icmlEqualContribution}

\begin{abstract}
For time series arising from latent dynamical systems, existing cross-domain generalization methods commonly assume that samples are comparably meaningful within a shared representation space. In real-world settings, however, different datasets often originate from structurally heterogeneous families of dynamical systems, leading to fundamentally distinct feature distributions. Under such circumstances, performing global alignment while neglecting structural differences is highly prone to establishing spurious correspondences and inducing negative transfer.
From the new perspective of cross-domain structural correspondence failure, we revisit this problem and propose a structurally stratified calibration framework. This approach explicitly distinguishes structurally consistent samples and performs amplitude calibration exclusively within structurally compatible sample clusters, thereby effectively alleviating generalization failures caused by structural incompatibility. Notably, the proposed framework achieves substantial performance improvements through a concise and computationally efficient calibration strategy.
Evaluations on 19 public datasets (100.3k samples) demonstrate that SSCF significantly outperforms strong baselines under the zero-shot setting. These results confirm that establishing structural consistency prior to alignment constitutes a more reliable and effective pathway for improving cross-domain generalization of time series governed by latent dynamical systems.
\end{abstract}

\section{Introduction}
Cross-dataset generalization is one of the key bottlenecks in deploying time-series models from controlled offline evaluations to open, real-world settings \cite{rasul2023lag}. Although models may achieve strong performance on source datasets, their performance often degrades consistently and substantially when transferred to unseen target datasets \cite{gao2024unitsunifiedmultitasktime}.

Existing domain generalization methods, such as DeepCORAL~\cite{sun2016deep}, MMD~\cite{gretton2012kernel}, DANN~\cite{ganin2016domain}, and CoDATS~\cite{liang2020we}, predominantly reduce domain discrepancies by constraining the distributions of representations across domains. At the modeling level, these approaches generally assume that samples from different domains are comparably meaningful in a shared feature space, while rarely making explicit when such comparability holds or how it should be assessed. Under this assumption, alignment or calibration operations are typically imposed uniformly at a global or class-wise granularity \cite{zhao2024mitigating, qin2024fdgnet}. In this work, we collectively refer to these strategies as Global Alignment, highlighting their common characteristic of enforcing uniform cross-domain alignment constraints without distinguishing sample-wise comparability.

However, for time series generated by latent dynamical systems, the assumption that cross-dataset samples are universally comparable within a single representation space does not always hold. Owing to heterogeneity in generation mechanisms, time series from different datasets may exhibit stable yet incompatible spectral shapes or spectral energy distributions \cite{guo2025enhancing}. Such differences are not necessarily attributable solely to devices or acquisition conditions, but may instead reflect structural disparities in the underlying dynamical processes \cite{de2008electroencephalographic, xie2024enhancing}. In this paper, we use the term structure to denote relatively stable spectral patterns in the frequency domain (e.g., peak locations, dominant energy bands, and their relative proportions) \cite{buckelmuller2006trait, tarokh2011trait}. These structures govern the relative geometric relationships among samples in spectral space, thereby constraining which cross-domain samples admit meaningful correspondences \cite{de2005electroencephalographic}.

\begin{figure}
    \centering
    \includegraphics[width=\linewidth]{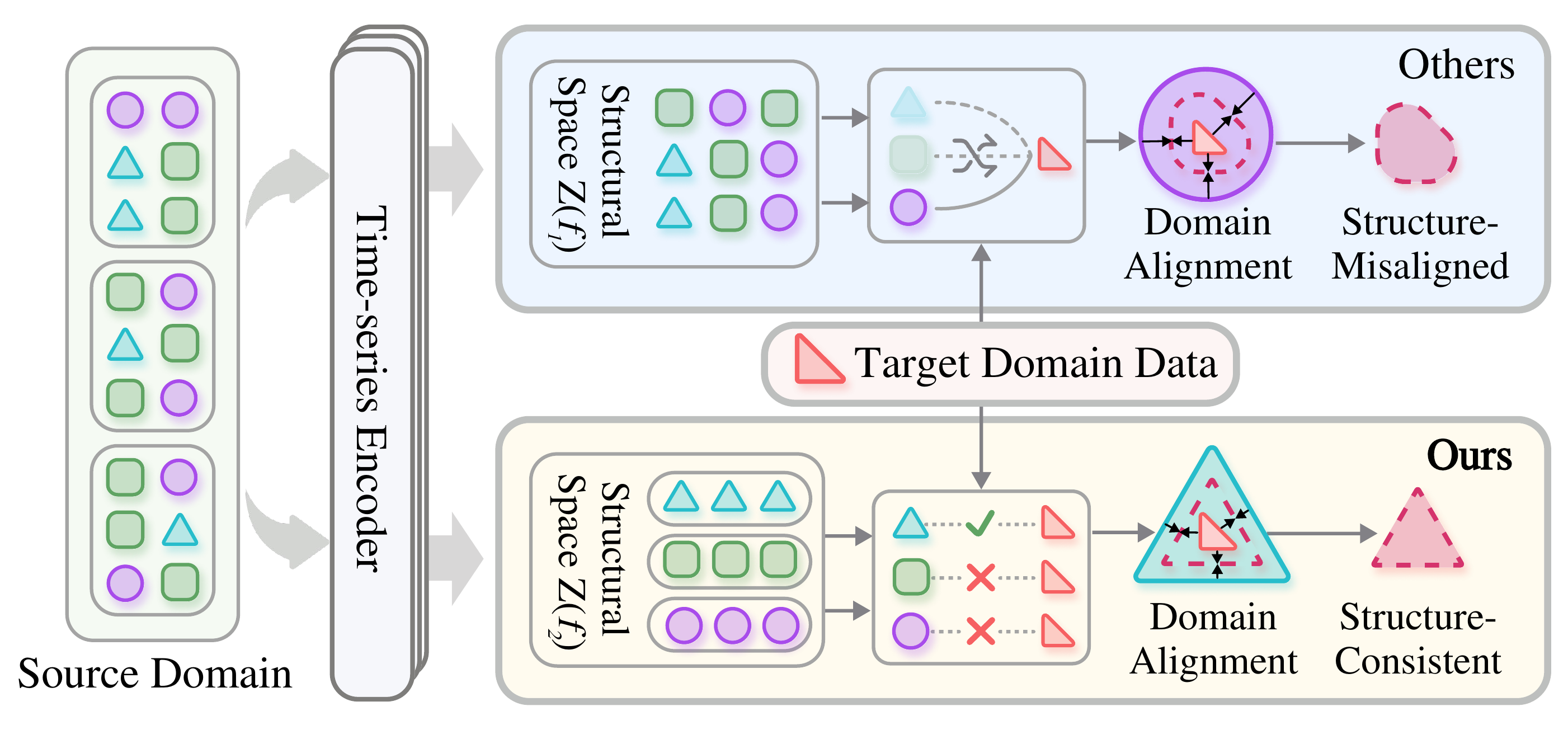}
    \caption{
    Overview of structure-consistent domain alignment for time-series domain generalization.
    (Other) Global alignment ignores structural heterogeneity, causing mismatches between structurally incompatible samples (dashed lines).
    (Our) We stratify samples by spectral structure and align only within consistent subsets, improving generalization stability.
    }
    \label{fig:overview}
\end{figure}

When datasets are structurally incompatible, even samples sharing identical semantic labels may lack stable one-to-one correspondences across domains \cite{zheng2024semantics}. In such cases, directly enforcing global alignment often introduces spurious correspondences between incompatible samples, thereby distorting structure-related representations and inducing negative transfer \cite{wang2024generalizable, liang2021pareto}. Intuitively, this situation is akin to aligning samples that are closer to a “triangle” onto a “circle” reference (Fig.~\ref{fig:overview}): the issue is not whether alignment is sufficiently strong, but whether the cross-domain correspondence itself is well defined \cite{yang2024spectral, gharib2025geometricmomentalignmentdomain}. Consequently, under structural heterogeneity without an explicit notion of comparability, global alignment can become ill-posed: multiple mutually incompatible correspondences may be equally “feasible” from an optimization perspective, even though some of them violate the constraints imposed by the underlying generative process \cite{luo2024geometric}. This observation suggests that calibration or alignment is better applied within sets of structurally comparable samples, where correspondences are more well-defined and interpretable \cite{pu2024unsupervised}.

These considerations indicate that the challenge of cross-dataset generalization lies not only in how strongly to align, but more fundamentally in whether alignment is grounded in identifiable structural correspondences. For such time series, structural comparability cannot be assumed a priori \cite{peng2024cross} ; therefore, a more principled prerequisite for alignment or calibration is to first identify structurally consistent and comparable subsets. Motivated by this insight, we explicitly introduce structural comparability assessment as a modeling component in cross-dataset generalization and organize subsequent calibration strategies accordingly.

Based on this perspective, we propose the \textbf{S}tructure-\textbf{S}tratified \textbf{C}alibration \textbf{F}ramework (SSCF), which adopts a “comparability-first, calibration-second” paradigm. SSCF first performs structure stratification to partition samples into multiple structurally compatible subsets; calibration is then carried out exclusively within each subset by constructing subset-specific references. This design reduces the spurious correspondences induced by enforcing alignment across structurally incompatible samples (as illustrated in Fig.~\ref{fig:overview}).

Our main contributions are summarized as follows:
\begin{enumerate}
    \item We characterize an applicability boundary for alignment/calibration in cross-dataset generalization: under structural heterogeneity without assessed comparability, directly applying global or class-level alignment is prone to introducing spurious cross-domain correspondences and incurring negative transfer.
    \item We propose the SSCF to operationalize structural comparability assessment: by first obtaining comparable subsets via structure stratification and then performing calibration within each subset, SSCF yields clearer calibration objectives and mitigates erroneous alignments in structurally heterogeneous settings.
    \item We validate the effectiveness of the proposed framework through zero-shot evaluation on 19 public datasets (100.3k samples in total), demonstrating that structure-conditioned calibration delivers more stable and consistent gains under multi-dataset transfer.
\end{enumerate}

\section{Related Work}

Cross-dataset generalization for time series is commonly studied under domain generalization (DG) and domain adaptation (DA), where models trained on one or multiple source datasets are expected to transfer to unseen domains.
Time-series data are particularly sensitive to acquisition devices, sensor configurations, and recording protocols, which can introduce notable distribution shifts even when semantic labels are aligned across datasets~\cite{ragab2023adatime}.
Beyond organizing prior work by technical tools, it is also useful to examine what is assumed to be comparable across domains.
This perspective clarifies when uniform alignment is well-defined and when it may induce negative transfer, motivating our focus on comparability assessment before calibration.

\subsection{Methods Assuming Global Cross-domain Comparability}

A broad family of DG/DA methods assumes that samples from different domains can be mapped into a shared representation space where geometric relations and similarity measures remain meaningful across domains.
Under this premise, cross-domain differences are treated as distribution shifts and mitigated via invariance or alignment objectives.
Representative examples include IRM~\cite{arjovsky2019invariant}, correlation alignment methods such as CORAL~\cite{sun2016deep}, discrepancy-based approaches (e.g., MMD), and adversarial alignment methods such as DANN and CDAN~\cite{ganin2016domain,long2018conditional}.
In time-series settings, many extensions add frequency-aware features, disentanglement constraints, or consistency regularization, while still enforcing alignment in a single shared latent space~\cite{chen2024cadt,guo2025frequencygraph}.
Despite their diversity, these methods implicitly rely on \emph{global comparability} in the learned space, so that uniformly coupling samples across domains is well-defined~\cite{lv2022causality}.

This assumption can be restrictive for time series generated by heterogeneous dynamical regimes.
When datasets correspond to distinct structural patterns, uniform alignment may introduce spurious correspondences and lead to negative transfer.
Our work addresses cross-dataset transferability from a complementary angle: before alignment or calibration, we first identify which subsets exhibit compatible structure and should therefore be processed together.

\subsection{Mechanism-aware and Disentangled Representation Learning}

Another line of work seeks transferable representations by separating task-relevant mechanisms from domain-specific factors, often motivated by causal or latent generative perspectives.
For example, CauDiTS~\cite{lu2024caudits} disentangles transferable mechanisms and domain-specific components from a causal viewpoint; PhASER~\cite{mohapatra2025phasedrivendomaingeneralizablelearning} emphasizes phase-related information; and related approaches leverage latent generative mechanisms or weak supervision to capture stable cross-domain structure~\cite{li2024time,cheng2023weakly}.
Generative modeling has also been used to factorize domain and task information in latent spaces, e.g., via hierarchical variational formulations or diffusion-based modeling~\cite{ilse2020diva,sun2021hierarchicalvariationalautoencodingunsupervised,zhang2025diffusion,ozyurt2023contrastivelearningunsuperviseddomain}.
Despite differences in modeling choices, these methods typically define invariance, disentanglement, or factorization objectives within latent representations, thereby presuming that samples across domains remain meaningfully comparable after representation learning~\cite{he2023domain,ding2024deep,Ismail_Fawaz_2025,chen2024disentangling}.
In contrast, SSCF emphasizes \emph{comparability selection}: rather than enforcing a single shared geometry across all samples, it restricts calibration to subsets that exhibit compatible shallow spectral characteristics.

\subsection{Spectral Representations and Frequency-domain Transfer}

Spectral representations are classical tools in time-series analysis for characterizing oscillatory patterns, rhythmic structure, and energy distributions.
In cross-domain transfer, Fourier-based methods such as FDA and its variants perform amplitude-driven style transfer by exchanging frequency-domain amplitude components while preserving phase information~\cite{yang2020fda,huang2021fsdr}.
Related ideas have also been introduced into time-series tasks, including multi-scale spectral modeling approaches based on wavelet transforms, empirical mode decomposition, federated time-series modeling, and Fourier-based calibration, as well as global amplitude normalization strategies to mitigate scale discrepancies across different datasets~\cite{zhang2025data,wang2022contrastive,fan2024domain,sarafraz2024domain,zhao2022test,wang2024generalizable}.
These approaches are effective when jointly processed samples share sufficiently consistent spectral structure.

However, time series often contain multiple dynamical regimes associated with distinct spectral characteristics.
In such cases, applying the same spectral manipulation to all samples may mix incompatible subspaces and weaken transferability~\cite{liu2022non,zhang2022survey}.
SSCF uses spectral representations in a different role: it first performs coarse-grained stratification using shallow spectral patterns and then conducts amplitude calibration exclusively within each stratum, making calibration conditional on structural compatibility.

\section{Preliminaries}
\paragraph{Notation.}
Let $T$ denote the length of a discrete-time signal, and let the observation space be $\mathcal{X}\subset\mathbb{R}^{T}$.
Given a learnable feature extraction operator $f_{\text{enc}}(\cdot)$, the raw signal is mapped to a multi-channel, one-dimensional feature map. We denote the feature space by $\mathcal{H}\subset\mathbb{R}^{C\times L}$, where $C$ is the number of channels and $L$ is the temporal length of the feature map. We consider $D$ data domains (datasets), indexed by $d \in \{1, \dots, D\}$. Each domain is associated with a data-generating distribution $\mathcal{P}_d$. Samples from domain $d$ are denoted as $(x^{(d)}, y^{(d)})\sim\mathcal{P}_d$, where
$x^{(d)} \in \mathcal{X}$ and $y^{(d)} \in \{1, \dots, C_y\}$ represents the class label. For a feature map $h\in\mathcal{H}$, we denote its frequency-domain representation as $H(f)$, which characterizes the response of each channel across different frequencies.

\paragraph{Cross-domain Setting.}
Under the cross-dataset generalization setting, we assume access to labeled samples from $D_s$ source domains$\{\mathcal{P}_d\}_{d=1}^{D_s}$ for model training, while the target domain $\mathcal{P}_t$ is entirely unseen during training. Owing to differences in acquisition devices, observation pipelines, or recording protocols, different domains may exhibit systematic distribution shifts in the feature space.

\paragraph{Problem Statement.}
Given source-domain samples $\{(x^{(d)}, y^{(d)})\}_{d=1}^{D_s}$, our objective is to learn a feature encoder
$f_{\mathrm{enc}} : \mathcal{X} \to \mathcal{H}$ together with a classifier $g : \mathcal{H} \to \{1, \dots, C_y\}$,
such that the resulting predictor $g \circ f_{\mathrm{enc}}$ achieves stable generalization performance on an unseen target domain $\mathcal{P}_t$.
Formally, we aim to minimize the target-domain risk
\begin{equation}
\mathcal{R}_t(f_{\mathrm{enc}}, g)
=
\mathbb{E}_{(x,y)\sim\mathcal{P}_t}
\big[\ell\big(g(f_{\mathrm{enc}}(x)),y\big)\big]
\end{equation}
while restricting the training process to source-domain data only, without accessing any target-domain samples for model training or hyperparameter selection.

\begin{figure*}
    \centering
    \includegraphics[width=1\linewidth]{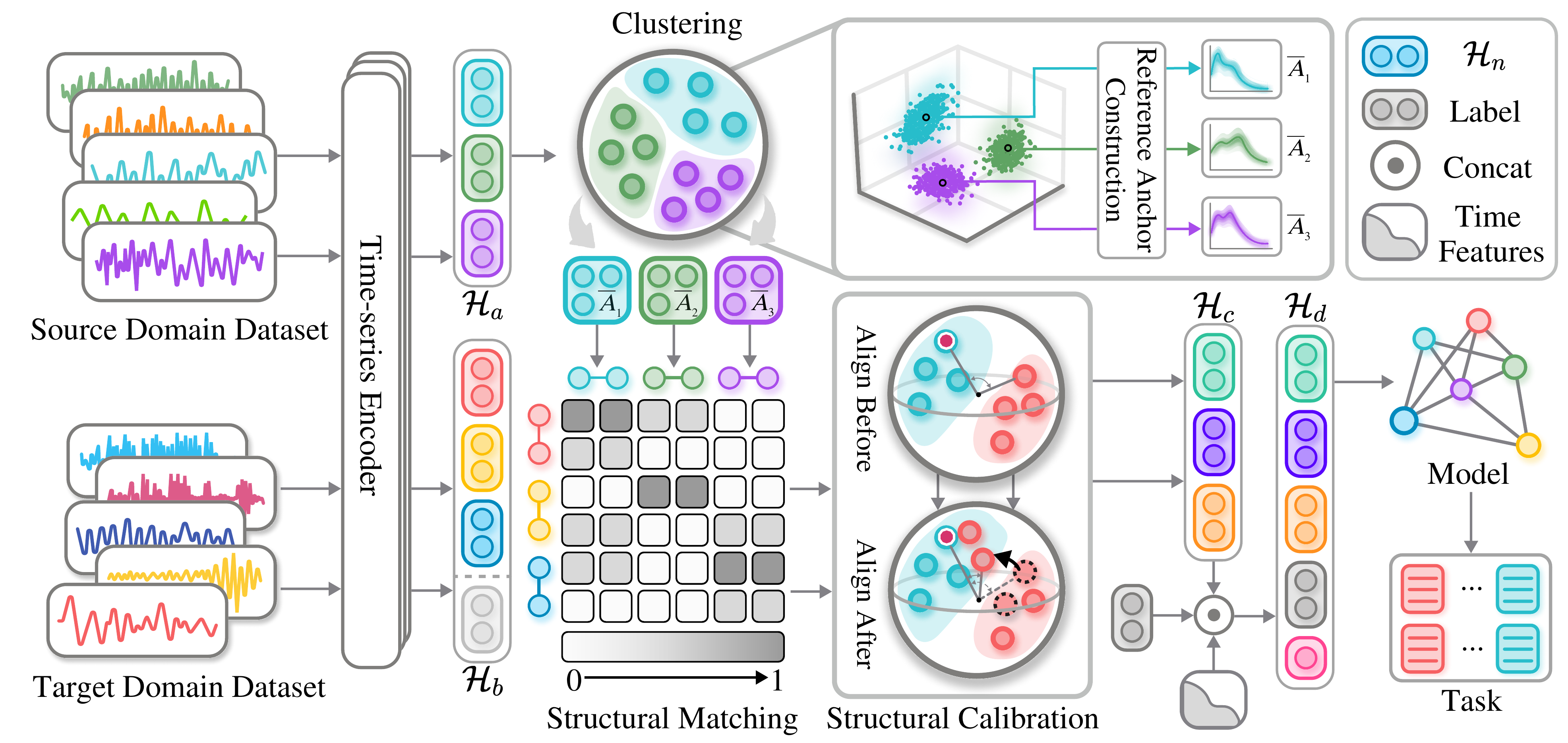}
    \caption{Overview of the proposed SSCF. Time-series data from the source and target domains are first encoded into shallow feature representations and subsequently stratified in the power spectral feature space to capture coarse-grained spectral patterns.
Within each structural stratum, a corresponding reference anchor is constructed.
For an arbitrary input sample, SSCF first performs structural matching to identify the most compatible stratum and then conducts amplitude calibration exclusively within that stratum, while explicitly preserving phase information. }
    \label{fig:framework}
\end{figure*}

\section{Method}
Most existing cross-dataset methods impose uniform alignment or invariance constraints across all samples, implicitly assuming that cross-domain samples can be consistently compared under a single metric in the representation space. We propose the SSCF, which first performs coarse-grained stratification of samples based on shallow spectral patterns, and then constructs reference anchors and conducts amplitude calibration exclusively within structurally consistent subsets (see Fig.~\ref{fig:framework}).

\subsection{Spectral Modeling Assumptions}

\paragraph{Definition 1 (Time Series from Latent Dynamical Systems).}
We consider time series generated from latent dynamical systems, where observations can be viewed as structure-related signals transformed by domain-specific observation processes. Under the cross-dataset setting, inter-domain differences may manifest not only in statistical distributions, but also in the channel-wise power spectral shapes and relative energy distributions of shallow features.

\paragraph{Definition 2 (Feature-level Spectral Observation Model).}
Let a sample $x^{(d)}$ be encoded into a shallow feature map
\begin{equation}
h^{(d)} = f_{\text{enc}}(x^{(d)})\in\mathbb{R}^{C\times L}
\end{equation}
Denote its complex frequency-domain representation by $H^{(d)}(f)$. We adopt the following empirical approximation for feature-level spectral decomposition:
\begin{equation}
H^{(d)}(f) \approx g_d(f)\odot Z(f)
\label{eq:feature_freq_model}
\end{equation}
where $Z(f)$ represents a structure-related latent spectral pattern that is relatively stable across domains, and $g_d(f)$ characterizes domain-specific, channel-wise frequency responses (amplitude/scale effects). This formulation serves as an empirical approximation to guide the design of stratification and calibration, rather than a strict generative model.

\subsection{Structure Stratification and Structural Representation}
Calibration or alignment is more appropriate between structurally compatible samples. To explicitly characterize structural differences across samples, we construct spectral pattern representations in the shallow feature space and perform coarse-grained stratification accordingly. The goal here is not to precisely recover the true underlying dynamical structure, but to obtain a partition that sufficiently separates major spectral pattern differences for subsequent within-structure calibration.

Given an input sample $x$, the encoder produces a shallow feature map
\begin{equation}
h = f_{\mathrm{enc}}(x)\in\mathbb{R}^{C\times L}
\end{equation}
where $C$ is the number of channels and $L$ is the spatial (temporal) length. After mean removal, we estimate the channel-wise power spectra using the Welch method:
\begin{equation}
P(x)=\mathcal{S}(h)\in\mathbb{R}_{+}^{C\times F}
\end{equation}
where $\mathcal{S}(\cdot)$ denotes the power spectral estimation operator and $F$ is the frequency dimension. We use $P(x)$ as the structural (spectral-pattern) representation of the sample.

In this representation space, we perform hard clustering using only source-domain samples $\{P(x)\}$ to obtain $K$ structurally compatible clusters:
\begin{equation}
\{\mathcal{C}_k\}_{k=1}^{K}
\end{equation}
where each cluster $\mathcal{C}_k$ denotes a subset of source-domain samples whose power spectral representations $P(x)$ share similar spectral patterns.

We adopt K-Means as the default implementation due to its simplicity and stability, which are sufficient for producing coarse-grained partitions. SSCF does not rely on precise structural recovery; the subsequent calibration only requires a stratification that separates the major spectral pattern differences.

\subsection{Reference Anchor Construction}

After stratification, samples within the same structural stratum may still exhibit amplitude or scale variations. To address this, we construct a mean-amplitude-squared (MAS) spectrum template for each structural stratum, which serves as a reference anchor representing a typical frequency-wise energy profile.

We define the MAS-based spectral center of each stratum to serve as its template. Specifically, suppose stratum $k$ contains $N_k = |\mathcal{C}_k|$ source-domain samples with channel-wise power spectra
\begin{equation}
P_i(c,f), \quad i=1,\dots,N_k,
\end{equation}
where $c$ indexes channels and $f$ indexes frequencies. We first take the square root of power to obtain amplitude spectra and then construct the anchor by squaring the mean amplitude:
\begin{equation}
\bar{P}_k(c,f)
=
\left(
\frac{1}{N_k}
\sum_{i=1}^{N_k}
\sqrt{P_i(c,f) + \epsilon}
\right)^2
\label{eq:geometric_center}
\end{equation}
where $\epsilon$ is a numerical stability constant. Eq.~(\ref{eq:geometric_center}) computes the square of the mean amplitude rather than an RMS quantity, yielding a mean-amplitude-squared (MAS) template. Compared with directly averaging power, this construction aggregates amplitudes before squaring, which makes the anchor less sensitive to rare large-power outliers. We use $\bar{P}_k$ as the reference anchor for stratum $k$, constructed once after encoder warm-up and then fixed for the remaining training and inference stages.

\begin{table*}[t]
\small
\centering
\caption{Overall performance overview (\%) for sleep staging under the Leave-One-Domain-Out (LODO) protocol and external target-domain evaluation.
 The LODO results report the average performance across five source-domain splits, while the external results are evaluated on six unseen target datasets.
 All results are reported as the mean $\pm$ standard deviation over five independent runs. The best results are highlighted in bold.}
\label{tab:sleep_results}
\setlength{\tabcolsep}{4.0pt}
\begin{tabular}{lcccccccc}
\toprule
\multirow{2}{*}{\textbf{Method}}
& \multicolumn{1}{c}{\textbf{LODO}}
& \multicolumn{7}{c}{\textbf{External Target Domains}} \\
\cmidrule(lr){2-2} \cmidrule(lr){3-9}
& \textbf{Avg}
& \textbf{ABC}
& \textbf{CCSHS}
& \textbf{CFS}
& \textbf{HMC}
& \textbf{ISRUC}
& \textbf{Sleep-EDFx}& \textbf{Avg} \\
\midrule

Baseline
& 59.76 $\pm$ 0.64
& 69.26 $\pm$ 1.32
& 75.60 $\pm$ 0.82
& 73.33 $\pm$ 0.71
& 65.08 $\pm$ 2.00
& 67.76 $\pm$ 1.36
& 70.55 $\pm$ 1.58
& 70.26 $\pm$ 1.30 \\

LG-SleepNet
& 57.45 $\pm$ 0.04
& 67.82 $\pm$ 0.13
& 67.68 $\pm$ 0.07
& 70.05 $\pm$ 0.06
& 65.18 $\pm$ 0.13
& 68.83 $\pm$ 0.14
& 57.14 $\pm$ 0.11
& 66.12 $\pm$ 0.11 \\

SleepEEGNet
& 53.22 $\pm$ 0.04
& 61.08 $\pm$ 0.13
& 58.11 $\pm$ 0.06
& 60.92 $\pm$ 0.05
& 61.97 $\pm$ 0.13
& 67.72 $\pm$ 0.13
& 57.07 $\pm$ 0.10
& 61.15 $\pm$ 0.10 \\

\midrule
IRM
& 56.46 $\pm$ 0.63
& 66.21 $\pm$ 1.53
& 69.36 $\pm$ 0.83
& 69.46 $\pm$ 0.73
& 62.72 $\pm$ 2.06
& 66.37 $\pm$ 1.45
& 65.80 $\pm$ 1.72
& 66.65 $\pm$ 1.39 \\

MMD
& 62.77 $\pm$ 0.58
& 70.30 $\pm$ 1.34
& 76.06 $\pm$ 0.78
& 74.18 $\pm$ 0.70
& 66.79 $\pm$ 1.93
& 69.11 $\pm$ 1.31
& 70.51 $\pm$ 1.60
& 71.16 $\pm$ 1.28 \\

CORAL
& 59.58 $\pm$ 0.61
& 69.81 $\pm$ 1.33
& 77.00 $\pm$ 0.76
& 74.78 $\pm$ 0.67
& 67.05 $\pm$ 1.97
& 69.00 $\pm$ 1.33
& 70.35 $\pm$ 1.68
& 71.33 $\pm$ 1.29 \\

SleepDG
& 60.92 $\pm$ 0.59
& 69.45 $\pm$ 1.29
& 77.33 $\pm$ 0.73
& 74.86 $\pm$ 0.66
& 65.89 $\pm$ 1.97
& 68.14 $\pm$ 1.32
& 70.33 $\pm$ 1.61
& 71.00 $\pm$ 1.26 \\

\midrule
\textbf{Ours}
& \textbf{69.37 $\pm$ 0.49}
& \textbf{71.96 $\pm$ 1.24}
& \textbf{81.04 $\pm$ 0.58}
& \textbf{77.12 $\pm$ 0.65}
& \textbf{72.49 $\pm$ 1.75}
& \textbf{74.76 $\pm$ 1.16}
& \textbf{73.33 $\pm$ 1.42}
& \textbf{75.12 $\pm$ 1.13} \\

\bottomrule
\end{tabular}
\end{table*}

\begin{table*}[t]
\small
\centering
\caption{Performance comparison (Macro-F1 score $\pm$ standard deviation) under the Leave-One-Domain-Out protocol.
The table is divided into two parts: Arrhythmia (top) and Human Activity Recognition (HAR) (bottom).
SD denotes the source domains, and TD denotes the target domain.
For each task, $K$ is selected solely based on source-domain validation, without access to any target-domain samples.
The best results are highlighted in bold.}
\label{tab:ecg_har_results}

\setlength{\tabcolsep}{9pt} 
\renewcommand{\arraystretch}{1.2} 

\begin{tabular}{l ccccc}
\toprule
\multicolumn{6}{c}{\textbf{Arrhythmia}} \\
\cmidrule(lr){1-6}

\textbf{SD} & \small II, III, IV & \small I, III, IV & \small I, II, IV & \small I, II, III &  \\
\textbf{TD} & \textbf{I} & \textbf{II} & \textbf{III} & \textbf{IV} & \textbf{Avg} \\
\cmidrule(lr){1-1} \cmidrule(lr){2-5} \cmidrule(lr){6-6}

Baseline & $59.30 \pm 0.21$ & $89.33 \pm 0.16$ & $83.98 \pm 0.16$ & $60.03 \pm 0.25$ & $73.16 \pm 0.20$ \\
CORAL    & $56.55 \pm 0.22$ & $\mathbf{89.59 \pm 0.15}$ & $86.73 \pm 0.15$ & $57.86 \pm 0.25$ & $72.68 \pm 0.19$ \\
MMD      & $56.71 \pm 0.21$ & $73.40 \pm 0.26$ & $60.45 \pm 0.27$ & $71.19 \pm 0.21$ & $65.44 \pm 0.24$ \\
IRM      & $55.17 \pm 0.22$ & $89.04 \pm 0.16$ & $88.92 \pm 0.14$ & $55.76 \pm 0.25$ & $72.22 \pm 0.19$ \\
\textbf{Ours}     & $\mathbf{69.56 \pm 0.17}$ & $89.51 \pm 0.16$ & $\mathbf{89.28 \pm 0.14}$ & $\mathbf{76.70 \pm 0.18}$ & $\mathbf{81.26 \pm 0.16}$ \\

\midrule[\heavyrulewidth] 
\addlinespace[0.5em]     

\multicolumn{6}{c}{\textbf{Human Activity Recognition}} \\
\cmidrule(lr){1-6}

\textbf{SD} & \small II, III, IV & \small I, III, IV & \small I, II, IV & \small I, II, III &  \\
\textbf{TD} & \textbf{I} & \textbf{II} & \textbf{III} & \textbf{IV} & \textbf{Avg} \\
\cmidrule(lr){1-1} \cmidrule(lr){2-5} \cmidrule(lr){6-6}

Baseline & $83.65 \pm 0.29$ & $82.74 \pm 0.46$ & $66.79 \pm 0.37$ & $98.09 \pm 0.05$ & $82.82 \pm 0.29$ \\
CORAL    & $84.49 \pm 0.28$ & $93.11 \pm 0.29$ & $81.23 \pm 0.27$ & $98.29 \pm 0.05$ & $89.28 \pm 0.22$ \\
MMD      & $70.39 \pm 0.39$ & $93.26 \pm 0.27$ & $73.35 \pm 0.34$ & $97.14 \pm 0.06$ & $83.54 \pm 0.27$ \\
IRM      & $92.35 \pm 0.18$ & $\mathbf{93.55 \pm 0.28}$ & $83.66 \pm 0.25$ & $97.74 \pm 0.06$ & $91.83 \pm 0.19$ \\
\textbf{Ours}     & $\mathbf{98.59 \pm 0.08}$ & $90.41 \pm 0.34$ & $\mathbf{89.14 \pm 0.20}$ & $\mathbf{99.56 \pm 0.02}$ & $\mathbf{94.43 \pm 0.16}$ \\

\bottomrule
\end{tabular}
\end{table*}

\subsection{Structural Matching and Intra-structural Amplitude Calibration}

Based on the above stratification and reference anchors, we perform structural matching for each input sample and conduct amplitude calibration within the matched stratum.
For an arbitrary input sample $x$, we first compute its structural representation $P(x)$
and identify the most compatible stratum among the set of structural centers
$\{\bar{P}_k\}_{k=1}^{K}$.
Since $\bar{P}_k$ is defined as a squared-mean power anchor, we perform matching in the corresponding square-rooted space to ensure dimensional consistency.
Specifically, we vectorize the power spectra across channel and frequency dimensions and measure the discrepancy using the Euclidean distance:
\begin{equation}
k^\ast(x)
=
\arg\min_{k}
\left\|
\mathrm{vec}\!\left(P(x)\right)
-
\mathrm{vec}\!\left(\sqrt{\bar{P}_k}\right)
\right\|_2
\label{eq:structure_matching}
\end{equation}
This rule corresponds to nearest-center matching in the joint channel–frequency power spectral space, providing each sample with a unique structural reference.

After matching to stratum $k^\ast(x)$, amplitude calibration is performed exclusively within that stratum. Let $H(c,f)$ denote the complex spectrum of the sample and $P(c,f)$ its power spectrum.
We preserve the phase and adjust the amplitude to match the square-rooted reference anchor:
\begin{equation}
\begin{aligned}
\min_{M} \quad
& \left\|
\mathrm{vec}\!\left(
\big| M(c,f)\odot H(c,f) \big|^2
-
\sqrt{\bar{P}_{k^\ast}(c,f)}
\right)
\right\|_2^2 \\
\text{s.t.} \quad
& M(c,f) \ge 0
\end{aligned}
\label{eq:amplitude_calibration}
\end{equation}
This objective corresponds to channel-wise and frequency-wise amplitude scaling, whose closed-form solution is
\begin{equation}
\tilde{H}(c,f)
=
M_{k^\ast}(c,f) \odot H(c,f)
\end{equation}
where the scaling factor is defined as
\begin{equation}
M_k(c,f)
=
\sqrt{
\frac{\sqrt{\bar{P}_k(c,f)}}
{P(c,f) + \epsilon}
}
\label{eq:spectral_transport}
\end{equation}
Eq.~(\ref{eq:amplitude_calibration}) provides a conceptual formulation of the desired amplitude adjustment, from which the closed-form scaling rule in Eq.~(\ref{eq:spectral_transport}) is obtained.
The calibrated feature representation is then obtained via inverse Fourier transform. Because calibration is restricted to the matched structural stratum, this procedure avoids inconsistent transformations induced by cross-structure mixing.

\subsection{Training protocol and calibration placement.}

SSCF follows a two-stage training protocol that separates structure estimation from discriminative learning.

\paragraph{Stage I: Structure estimation.}
We first train an auxiliary encoder using standard ERM on source domains. This model is used only to extract spectral representations for clustering and anchor construction.
The resulting anchors are fixed thereafter and the auxiliary encoder is discarded.
It does not share parameters with the final model.

\paragraph{Stage II: End-to-end training with fixed anchors.}
We then train the final encoder--classifier model from scratch.
For each input sample, encoder features are deterministically calibrated
using the precomputed anchors before classification.
The anchors are treated as constants and do not receive gradients.
The calibration operation is differentiable with respect to encoder outputs,
allowing gradients to propagate back to the encoder parameters.
Encoder and classifier are therefore jointly optimized under the task loss.

\section{Experiments}
\subsection{Experimental Setups}

\paragraph{Datasets and Evaluation Protocol.}
We evaluate our method on 19 public datasets spanning three task categories (100.3k samples in total) under the Leave-One-Domain-Out (LODO) protocol, where one dataset is held out as an unseen target domain and the remaining datasets are used as source domains for training.
For sleep staging, five datasets (MESA, MROS1, MROS2, SHHS1, P2018) are used as source domains, with zero-shot evaluation on six independent datasets (ABC, CCSHS, CFS, ISRUC, HMC, Sleep-EDFx).
For arrhythmia detection, four ECG datasets (MIT-BIH, INCART, PTB-XL, CHAPMAN) are evaluated.
For human activity recognition (HAR), four datasets (UCIHAR, MotionSense, WISDM, USCHAD) are evaluated under the same protocol.
Across all tasks, structure stratification and reference anchors are estimated using source-domain training data only, without any target-domain participation.

\paragraph{\textbf{Preprocessing.}}
Across all tasks, raw signals are segmented into fixed-length windows and normalized independently per dataset, without access to target-domain samples.
For sleep staging, PSG signals are segmented into 30-second epochs, band-pass filtered (0.3--35 Hz), resampled to 100 Hz, Z-score normalized, and labels are remapped by merging N3/N4 according to the AASM standard, following SleepDG\cite{wang2024generalizable} .
For arrhythmia and HAR, standard windowing and normalization are applied following prior work.

\paragraph{Implementation Details.}
Macro-F1 (MF1) is used as the primary evaluation metric.
Models are trained using Adam for 200 epochs with a learning rate of $1\times10^{-3}$ and a batch size of 128.
Each experiment is repeated with five random seeds, and results are reported as mean $\pm$ standard deviation.
SSCF strictly follows the domain generalization setting: structure stratification and reference anchors are fixed after training, and inference on unseen target samples involves structure assignment, within-stratum amplitude calibration, and discriminative prediction, without target-domain fine-tuning or test-time adaptation.

\subsection{Results and Analysis}
\paragraph{Sleep Staging.}
As shown in Table~\ref{tab:sleep_results}, non-DG methods exhibited relatively weak overall performance on external target domains. Introducing domain generalization constraints led to moderate improvements over non-DG baselines. In contrast, SSCF achieved higher average performance under both source-domain LODO evaluation and on most external target domains. We also observed that the performance gains varied across target domains, suggesting that the effectiveness of structure-stratified calibration depends on the degree of distributional or structural discrepancy between the source and target domains.

\paragraph{Arrhythmia Detection.}
Table~\ref{tab:ecg_har_results} reports the LODO results for arrhythmia detection. While most baseline methods attained reasonable average performance, they suffered pronounced degradation on certain target domains, reflecting the impact of cross-dataset discrepancies on model transferability. SSCF generally demonstrated advantages in both average performance and worst-domain performance, although the magnitude of improvement again varied across target domains.

\paragraph{Human Activity Recognition.}
On the human activity recognition task, SSCF also improved both average and worst-domain performance, as shown in Table~\ref{tab:ecg_har_results}. The gains were modest on some target domains, indicating that when sensor configurations and acquisition conditions differ substantially, the benefits of structure-stratified calibration still depend on the proximity between the structural distributions of the source and target domains.

\subsection{Ablation Study}
\begin{table*}
\centering
\caption{
Ablation study on alignment strategies (Macro-F1).
Results are reported under the leave-one-domain-out (LODO) evaluation protocol.} 
\label{tab:ablation_study}

\resizebox{\textwidth}{!}{
    \setlength{\tabcolsep}{3.5pt}
    \renewcommand{\arraystretch}{1.2}
    
    \begin{tabular}{l c cccccc c}
    \toprule
    \multirow{2}{*}{\textbf{Alignment Strategy}}
    & \multicolumn{1}{c}{\textbf{LODO}}
    & \multicolumn{7}{c}{\textbf{External Target Domains}} \\
    \cmidrule(lr){2-2} \cmidrule(lr){3-9}
    
    & \textbf{Avg}
    & \textbf{ABC}
    & \textbf{CCSHS}
    & \textbf{CFS}
    & \textbf{HMC}
    & \textbf{ISRUC}
    & \textbf{Sleep-EDFx}
    & \textbf{Avg} \\
    \midrule

    Baseline
    & 59.76 $\pm$ 0.64
    & 69.26 $\pm$ 1.32
    & 75.60 $\pm$ 0.82
    & 73.33 $\pm$ 0.71
    & 65.08 $\pm$ 2.00
    & 67.76 $\pm$ 1.36
    & 70.55 $\pm$ 1.58
    & 70.26 $\pm$ 1.30 \\

    Global Alignment
    & 66.40 $\pm$ 0.48
    & 67.94 $\pm$ 1.40
    & 76.65 $\pm$ 0.68
    & 73.63 $\pm$ 0.69
    & 66.94 $\pm$ 1.90
    & 68.72 $\pm$ 1.45
    & 72.81 $\pm$ 1.44
    & 71.12 $\pm$ 1.26 \\

    Dataset-based Anchor
    & 63.14 $\pm$ 0.48
    & \textbf{72.37 $\pm$ 1.27}
    & 78.79 $\pm$ 0.70
    & 75.61 $\pm$ 0.68
    & 69.98 $\pm$ 1.80
    & 72.73 $\pm$ 1.41
    & \textbf{74.10 $\pm$ 1.33}
    & 73.93 $\pm$ 1.20 \\

    \midrule
    
    \textbf{Ours (Structural)}
    & \textbf{69.37 $\pm$ 0.49}
    & 71.96 $\pm$ 1.24
    & \textbf{81.04 $\pm$ 0.58}
    & \textbf{77.12 $\pm$ 0.65}
    & \textbf{72.49 $\pm$ 1.75}
    & \textbf{74.76 $\pm$ 1.16}
    & 73.33 $\pm$ 1.42
    & \textbf{75.12 $\pm$ 1.13} \\

    \bottomrule
    \end{tabular}
}
\end{table*}

To analyze the contributions of individual components in SSCF, we conducted ablation experiments under the same LODO setting as the main experiments. The results are summarized in Table~\ref{tab:ablation_study}. Unless otherwise specified, the encoder, classifier, and training hyperparameters were kept unchanged, and the number of strata $K$ as well as the matching and calibration procedures were identical to those of the full method. The compared variants include:
\begin{itemize}
    \item \textbf{Baseline}: Discriminative training without spectral calibration;
    \item \textbf{Global Anchor Calibration}: Constructing a single reference anchor using all source-domain samples and applying uniform calibration to all samples;
    \item \textbf{Domain-wise Anchor Calibration}: Constructing dataset-specific anchors and performing calibration within each dataset;
    \item \textbf{SSCF (Ours)}: Performing structure stratification first, followed by anchor construction and calibration within each structural stratum.
\end{itemize}
As shown in the table, global anchor calibration led to performance degradation on several target domains, indicating that applying a single calibration to all samples without distinguishing structural comparability can introduce mismatches. Domain-wise anchor calibration alleviated this issue on some target domains but yielded unstable overall improvements. In contrast, SSCF achieved more consistent gains across most settings, demonstrating that the “stratify-first, calibrate-within-stratum” constraint plays a critical role in effective calibration.

\subsection{Sensitivity Analysis: Structural Granularity and Anchor Matching}

\paragraph{Effect of Structural Granularity ($K$).}
We investigated the effect of structural granularity on performance for sleep staging and human activity recognition, evaluating $K\in{1,\dots,8}$. The corresponding results are shown in the top two subplots of the figure. Overall, performance was relatively weak when $K=1$ (i.e., no structure stratification). When $K\geq 2$, performance improved markedly and stabilized for $K\geq 3$, with only minor fluctuations thereafter. These results indicate that SSCF does not rely on fine-grained structural partitioning; rather, coarse-grained stratification is sufficient to capture the dominant spectral pattern differences, consistent with the design motivation of using stratification as a coarse prior constraint.

\paragraph{Impact of Anchor Matching Quality ($R$).}
We further examined the relationship between anchor matching quality and performance, with results shown in the bottom two subplots of the figure. Let $R$ denote the matching rank based on geometric distance, ordered from closest to farthest, where $R{=}1$ corresponds to the nearest reference anchor. This analysis was conducted on both sleep staging and human activity recognition tasks. Overall, higher matching quality (smaller geometric distance) tended to correspond to better predictive performance, particularly for $R{=}1,2,3$, where the relationship was relatively consistent. However, for $R{\geq}4$, this monotonic trend no longer held: a smaller geometric distance did not necessarily yield better performance. This observation suggests that geometric distance serves only as an approximate measure of structural similarity, and that effective calibration ultimately depends on structural comparability constraints rather than distance-based ranking alone.

\begin{figure}[t]
    \centering
    \begin{subfigure}[t]{0.48\columnwidth}
        \centering
        \includegraphics[width=\linewidth]{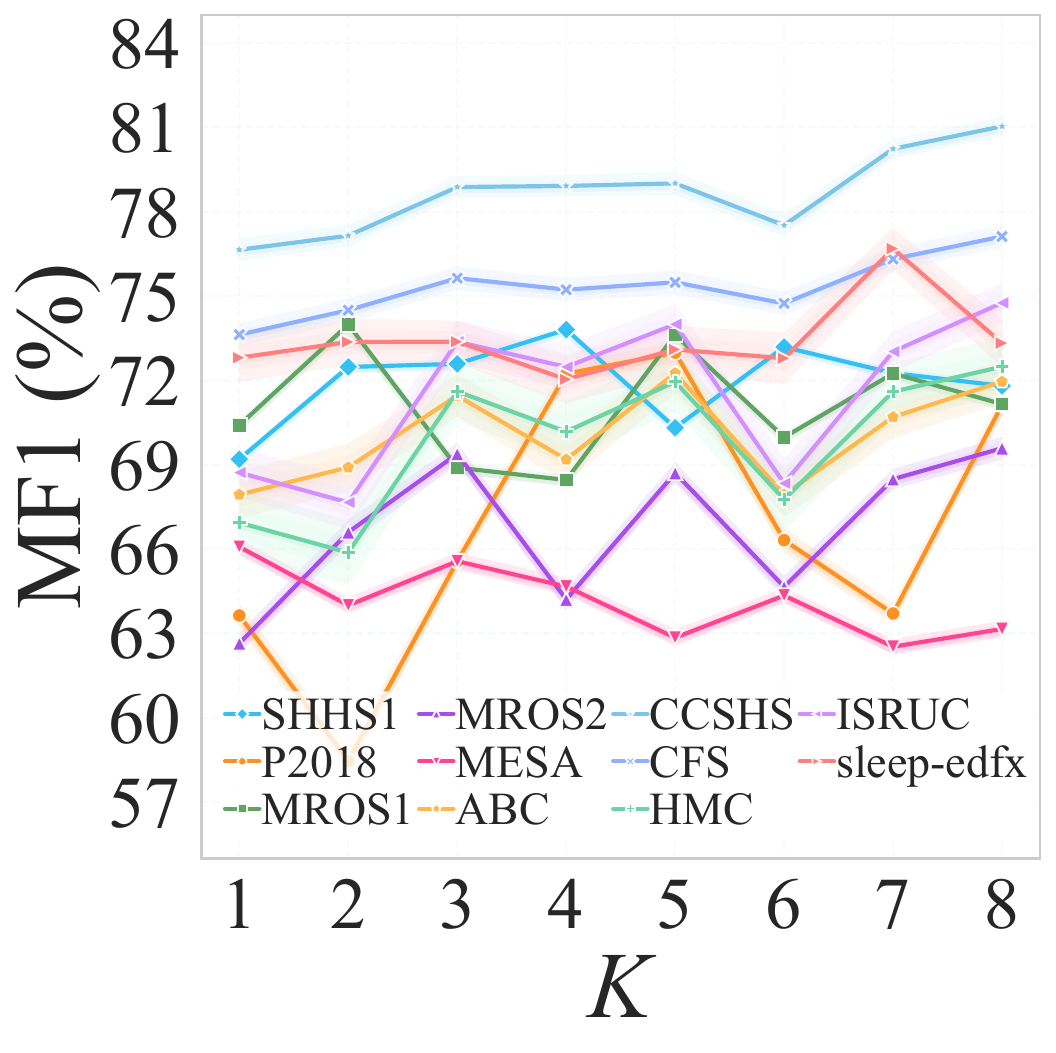}
        \caption{Sleep: varying $K$}
    \end{subfigure}
    \hfill
    \begin{subfigure}[t]{0.48\columnwidth}
        \centering
        \includegraphics[width=\linewidth]{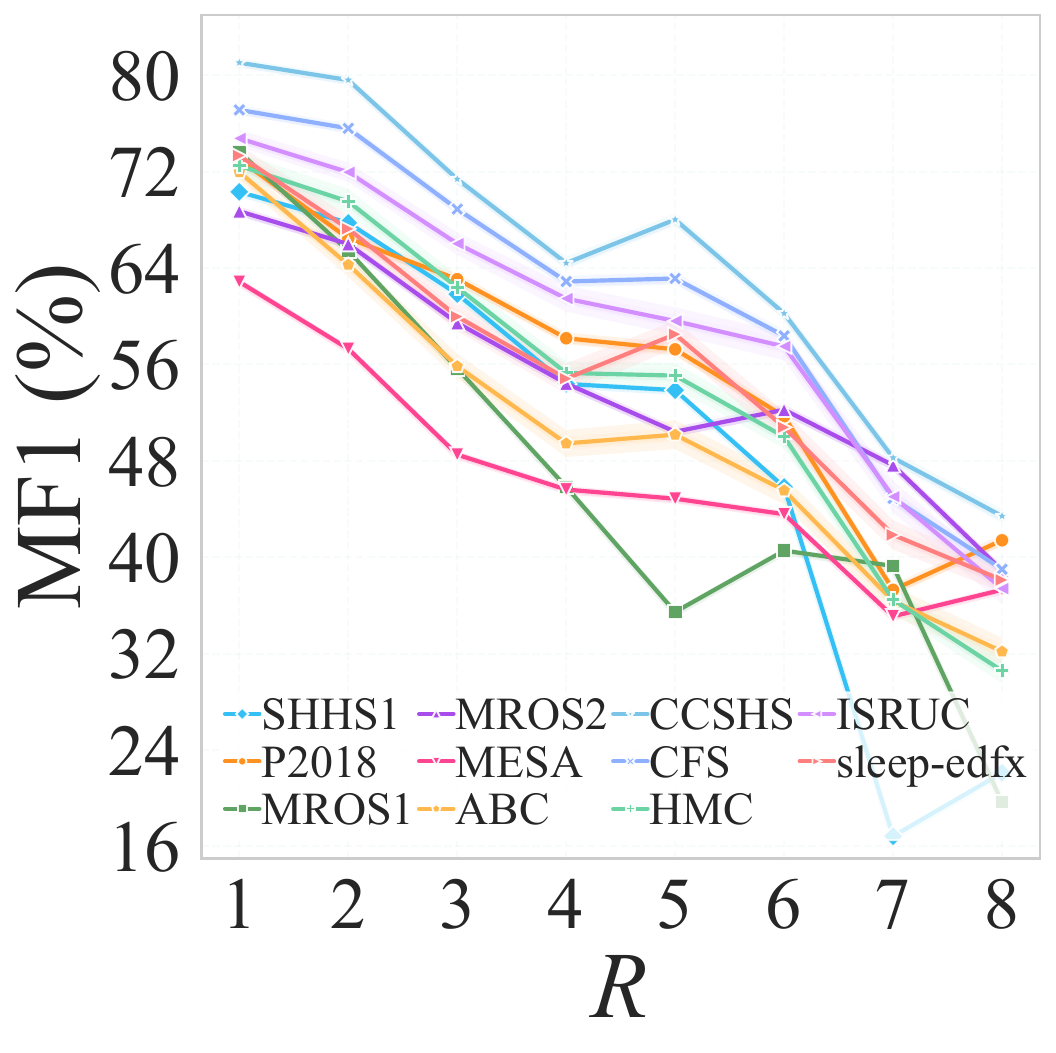}
        \caption{Sleep: rank $R$}
    \end{subfigure}

    \vspace{0.3em}

    \begin{subfigure}[t]{0.48\columnwidth}
        \centering
        \includegraphics[width=\linewidth]{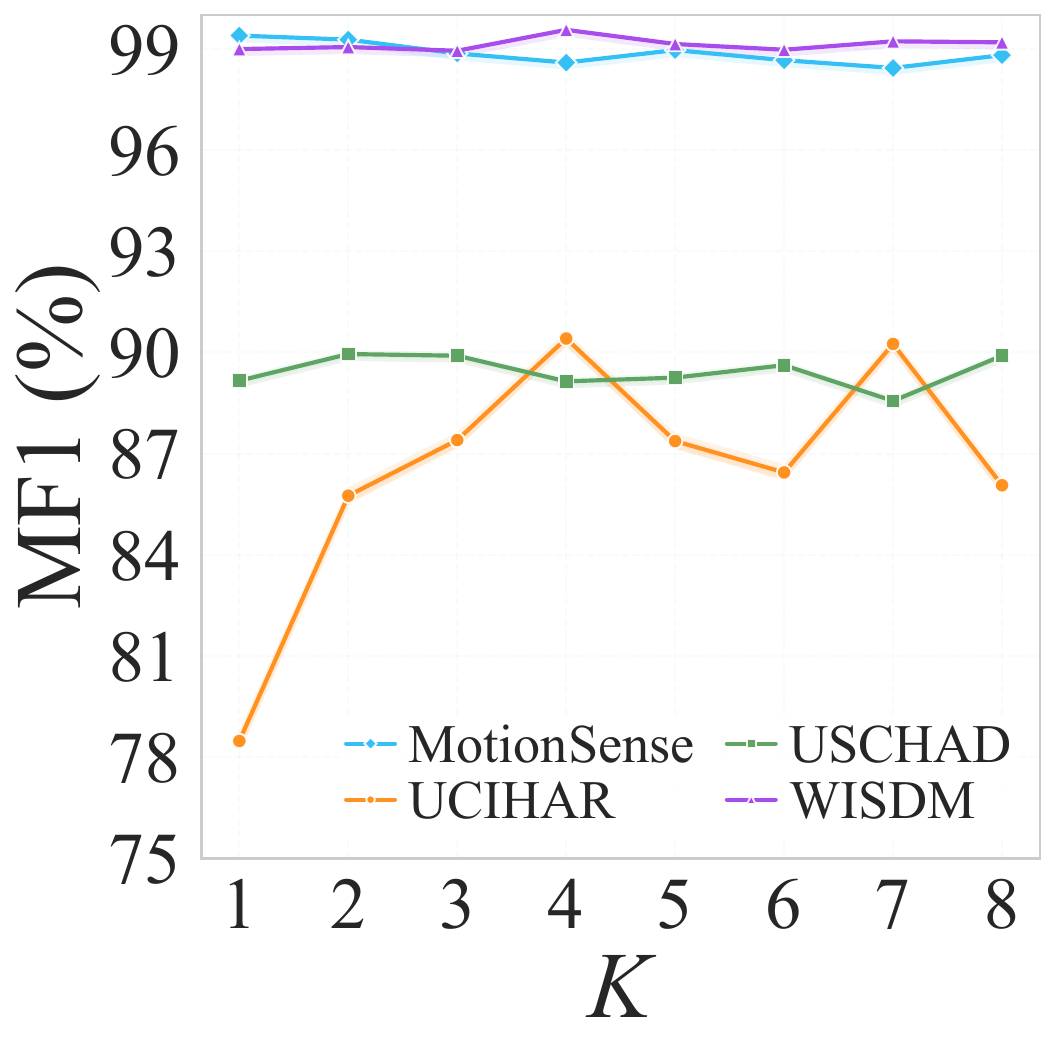}
        \caption{HAR: varying $K$}
    \end{subfigure}
    \hfill
    \begin{subfigure}[t]{0.48\columnwidth}
        \centering
        \includegraphics[width=\linewidth]{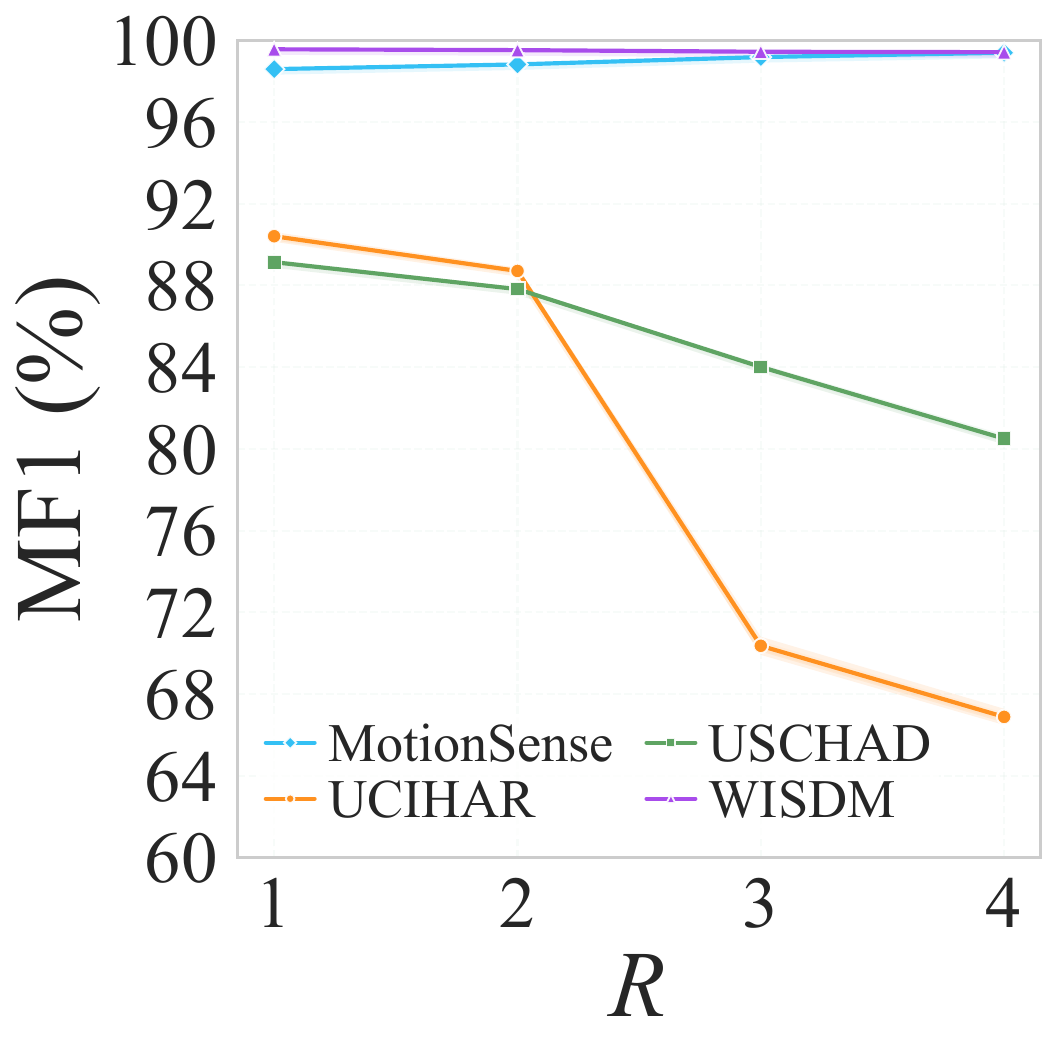}
        \caption{HAR: rank $R$}
    \end{subfigure}

    \caption{
    Sensitivity analysis of SSCF.
    Top: effect of structural granularity $K$.
    Bottom: impact of anchor matching rank $R$.
    }
    \label{fig:sensitivity_KR}
\end{figure}

\section{Conclusion}
This work revisits calibration in cross-dataset generalization from the perspective of structural heterogeneity in time series and proposes the SSCF. SSCF first performs structure stratification and then applies amplitude calibration based on reference anchors exclusively within structurally consistent subsets. We evaluate SSCF under a zero-shot setting on 19 public datasets. Experimental results demonstrate that SSCF generally yields more stable performance improvements across multiple tasks and target domains, although the magnitude of gains is not entirely uniform across domains. Future work will further investigate stratification and calibration strategies under non-stationary dynamics, where structural patterns evolve over time.

\section*{Impact Statement}
This paper presents work whose goal is to advance the field of Machine Learning. There are many potential societal consequences of our work, none which we feel must be specifically highlighted here.
\nocite{langley00}

\bibliography{main}
\bibliographystyle{icml2026}

\newpage
\appendix
\onecolumn
\section{Datasets Overview}
\subsection{Sleep Staging Datasets}
\begin{table}[htbp]
  \centering
  \caption{Detailed Overview of Sleep Staging Datasets and Signal Configuration}
  \label{tab:sleep_datasets_detail}
  \renewcommand{\arraystretch}{1.2}
  \setlength{\tabcolsep}{5pt}
  \begin{tabular}{l c c l l l l}
    \toprule
    \multirow{2}{*}{\textbf{Dataset}} & \multirow{2}{*}{\textbf{Subjects}} & \textbf{Fs} & \multicolumn{2}{c}{\textbf{EEG Configuration}} & \multicolumn{2}{c}{\textbf{EOG Configuration}} \\
    \cmidrule(lr){4-5} \cmidrule(lr){6-7}
    & & \textbf{(Hz)} & \textbf{Raw Channel} & \textbf{Ref} & \textbf{Raw Channel} & \textbf{Ref} \\
    \midrule

    \textbf{SHHS1}     & 5,652 & 125 & EEG & --- & EOG(L) & --- \\
    \textbf{P2018}     & 993   & 200 & C3-M2 & M2 & E1-M2 & M2 \\
    \textbf{MROS1}     & 2,872 & 256 & C4 & A1 & LOC & A2 \\
    \textbf{MROS2}     & 1,007 & 256 & C4 & M1 & E1 & M2 \\
    \textbf{MESA}      & 2,034 & 256 & EEG1 & (Pre-ref) & EOG-L & (Pre-ref) \\
    \textbf{ABC}       & 131   & 256 & F4 & M1 & E1 & M2 \\
    \textbf{CCSHS}     & 514   & 128 & C4 & A1 & LOC & A2 \\
    \textbf{CFS}       & 722   & 128 & C4 & M1 & LOC & M2 \\
    \textbf{HMC}       & 150   & 256 & EEG C4-M1 & M1 & EOG E1-M2 & M2 \\
    \textbf{ISRUC}     & 117   & 200 & F4-A1 & A1 & LOC-A2 & A2 \\
    \textbf{Sleep-EDFx} & 100  & 100 & EEG Fpz-Cz & Cz & EOG horizontal & --- \\

    \bottomrule
  \end{tabular}
  \footnotesize
  \flushleft
  \textbf{Note:} Fs = Sampling Rate; Ref = Reference Electrode. ‘---’ indicates that the specific reference information is not applicable or the signal is already pre-referenced in the provided dataset.
\end{table}

\subsection{ECG Arrhythmia Classification Datasets}
\begin{table}[htbp]
  \centering
  \caption{Overview of ECG Arrhythmia Classification Datasets Used in This Study.}
  \label{tab:ecg_datasets_detail}
  \renewcommand{\arraystretch}{1.2}
  \setlength{\tabcolsep}{5pt}
  \begin{tabular}{l c c c l}
    \toprule
    \textbf{Dataset} & \textbf{Subjects} & \textbf{Fs (Hz)} & \textbf{Original Leads} & \textbf{Used Lead} \\
    \midrule
    \textbf{PTB-XL}    & 18,885 & 500 & 12 (Standard) & Lead II \\
    \textbf{CHAPMAN}   & 10,646 & 500 & 12 (Standard) & Lead II \\
    \textbf{INCART}    & 75     & 257 & 12 (Standard) & Lead II \\
    \textbf{MITBIHADB} & 48     & 360 & 2 (MLII, V5)  & MLII \\
    \bottomrule
  \end{tabular}
  \footnotesize
  \flushleft
\end{table}

\subsection{Human Activity Recognition Datasets}
\begin{table}[htbp]
  \centering
  \caption{Overview of Human Activity Recognition (HAR) Datasets.}
  \label{tab:har_datasets_detail}
  \renewcommand{\arraystretch}{1.2}
  \setlength{\tabcolsep}{5pt}
  \begin{tabular}{l c c l l}
    \toprule
    \textbf{Dataset} & \textbf{Subjects} & \textbf{Fs (Hz)} & \textbf{Device Position} & \textbf{Selected Sensors} \\
    \midrule
    \textbf{WISDM}       & 51 & 20  & Waist / Pocket & Accelerometer \\
    \textbf{UCIHAR}      & 30 & 50  & Waist          & Accelerometer \\
    \textbf{MotionSense} & 24 & 50  & Pocket         & Accelerometer \\
    \textbf{USCHAD}      & 14 & 100 & Pocket         & Accelerometer \\
    \bottomrule
  \end{tabular}
  \footnotesize
  \flushleft
  \textbf{Note:} Fs denotes the sampling frequency. Although some datasets contain gyroscope data, only tri-axial accelerometer signals were used.
\end{table}
\clearpage

\section{Baseline Backbone Architectures}
All domain generalization methods and our proposed approach share the same backbone architecture within each task.

\subsection{Sleep Staging}
For sleep staging, we employ a hierarchical CNN--Transformer backbone designed to process raw EEG and EOG signals.

Each 30-second epoch is first encoded by a multi-layer 1D convolutional network, consisting of convolution, batch normalization, and ReLU activation blocks. The CNN produces a fixed-dimensional embedding for each epoch. The sequence of epoch embeddings is then processed by a single-layer Transformer encoder to model temporal dependencies across consecutive epochs. The output representations are passed to a linear classifier for sleep stage prediction.

\subsection{ECG Arrhythmia Classification}
For ECG-based arrhythmia classification, we adopt a CNN--BiLSTM backbone commonly used in physiological time-series analysis.

Raw ECG signals are first processed by a multi-stage 1D convolutional feature extractor, where each stage consists of convolution, normalization, non-linear activation,
and pooling operations to capture local morphological patterns. The resulting feature sequences are fed into a two-layer bidirectional LSTM (BiLSTM) to model temporal dependencies. Temporal max-pooling is applied over the BiLSTM outputs, followed by a multilayer perceptron (MLP) classifier.

\subsection{Human Activity Recognition}
For human activity recognition based on wearable sensor data, we employ the same CNN--BiLSTM backbone architecture as in the ECG task.

Although HAR signals are non-physiological in nature,
they exhibit similar temporal characteristics such as local pattern composition and sequential dependency. Using an identical backbone allows for controlled comparison of learning strategies without introducing task-specific architectural tuning.
\clearpage

\subsection{Offline MAS Template Construction}

Structural anchors are constructed only once using source-domain data.
We first pretrain the encoder using standard ERM to obtain stable shallow feature representations, after which the encoder parameters are frozen.
For each source-domain sample, we compute the channel-wise power spectral density (PSD) of its shallow features using the Welch method, and perform hard clustering via K-Means in the power-spectral space to obtain $K$ structural clusters.

For each structural cluster, we compute its MAS template in the amplitude domain and square it to obtain the corresponding power-domain reference anchor $\bar{P}_k$.
All structural clusters and reference anchors are kept fixed during subsequent training and inference, and do not participate in parameter updates.

\begin{algorithm}
\caption{Offline: Compute MAS templates (Reference Anchors) from Source Domains}
\label{alg:sscf_offline_centers}
\begin{algorithmic}[1]

\REQUIRE Source-domain dataset $\mathcal{D}_s=\{(x,y)\}$, encoder $f_{\text{enc}}(\cdot)$, window function $w$, frame length $F$, hop size $hop$, number of clusters $K$, warm-up epochs $N_0$, stability constant $\epsilon$

\ENSURE Structural clusters $\{\mathcal{C}_k\}_{k=1}^K$, MAS templates $\{A_k\}_{k=1}^K$, reference anchors (power) $\{\bar{P}_k\}_{k=1}^K$

\STATE \textbf{Stage 1: Warm-up for Shallow Features}
\STATE Train $(f_{\text{enc}}, g)$ using standard ERM for $N_0$ epochs (details omitted)
\STATE Freeze encoder parameters: $\textsc{Freeze}(f_{\text{enc}})$

\STATE \textbf{Stage 2: Spectral Encoding on Source Domains (Welch PSD)}
\STATE Initialize spectral set: $\mathcal{P}\leftarrow \emptyset$
\FOR{each source sample $x \in \mathcal{D}_s$}
    \STATE $h \leftarrow f_{\text{enc}}(x)$ \COMMENT{$h\in\mathbb{R}^{C\times L}$}
    \STATE Mean removal: $h \leftarrow h-\text{mean}(h,\text{dim}=L)$
    \STATE Framing and windowing: $h_{\text{frames}} \leftarrow \textsc{Frame}(h,F,hop)\odot w$
    \STATE PSD (Welch): $P(x)\leftarrow \mathbb{E}_{t}\big[\,|\text{FFT}(h_{\text{frames}}[t])|^2\,\big]$
    \STATE $\mathcal{P}\leftarrow \mathcal{P}\cup\{P(x)\}$
\ENDFOR

\STATE \textbf{Stage 3: Structural Stratification (K-Means in Power-Spectral Space)}
\STATE $\{\mathcal{C}_k\}_{k=1}^{K}\leftarrow \textsc{KMeans}(\mathcal{P},K)$

\STATE \textbf{Stage 4: MAS Template Construction (Amplitude Domain)}
\FOR{each cluster $k \in [1,K]$}
    \STATE $\{P_i\}_{i=1}^{N_k} \leftarrow \{P(x)\,|\,x\in\mathcal{C}_k\}$
    \STATE \textbf{MAS template (amplitude):} $A_k \leftarrow \frac{1}{N_k}\sum_{i=1}^{N_k}\sqrt{P_i+\epsilon}$
    \STATE \textbf{Reference anchor (power):} $\bar{P}_k \leftarrow A_k^2$
\ENDFOR

\STATE \textbf{return} $\{\mathcal{C}_k\}_{k=1}^K,\ \{A_k\}_{k=1}^K,\ \{\bar{P}_k\}_{k=1}^K$

\end{algorithmic}
\end{algorithm}

\subsection{Online Structural Matching and Calibration}

During both training and testing, a Welch-style PSD descriptor is computed for each input sample and matched to all reference anchors using Euclidean distance in the joint channel--frequency space.
The nearest structural anchor is selected as the reference.

Amplitude calibration is then performed \emph{only within the matched structural stratum}: the spectral phase is preserved, while the amplitude is scaled in a channel-wise and frequency-wise manner so that the calibrated spectrum matches the corresponding reference anchor.
In practice, we compute the scaling mask from Welch PSD descriptors and apply it to the full-length FFT magnitude as a fixed, non-parametric transformation.
The calibration operator is detached from backpropagation.

\begin{algorithm}[H]
\caption{Online: Structural Matching and Phase-Preserving Amplitude Calibration}
\label{alg:sscf_online_infer}
\begin{algorithmic}[1]

\REQUIRE Input features $X \in \mathbb{R}^{B\times C\times L}$, window function $w$, frame length $F$, hop size $hop$, reference anchors $\{\bar{P}_k\}_{k=1}^{K}$, stability constant $\epsilon$

\ENSURE Calibrated features $Y \in \mathbb{R}^{B\times C\times L}$

\STATE \textbf{Stage 1: Spectral Descriptor (Welch PSD, for Matching Only)}
\STATE Mean removal: $X \leftarrow X-\text{mean}(X,\text{dim}=L)$
\STATE Framing and windowing: $X_{\text{frames}} \leftarrow \textsc{Frame}(X,F,hop)\odot w$
\STATE PSD descriptor (Welch): $P_{\text{src}}^{\text{welch}} \leftarrow \mathbb{E}_{t}\big[\,|\text{FFT}(X_{\text{frames}}[t])|^2\,\big]$

\STATE \textbf{Stage 2: Structural Matching (Nearest-Anchor in PSD Space)}
\FOR{each sample $i \in [1,B]$}
    \STATE $d_{i,k}\leftarrow \|\mathrm{vec}((P_{\text{src}}^{\text{welch}})^{(i)})-\mathrm{vec}(\bar{P}_k)\|_2,\ \forall k\in[1,K]$
    \STATE $k^\ast \leftarrow \arg\min_k d_{i,k}$
    \STATE $P_{\text{ref}}^{(i)} \leftarrow \bar{P}_{k^\ast}$
\ENDFOR

\STATE \textbf{Stage 3: Amplitude Calibration}
\STATE \textbf{Stop gradient:} $\textsc{StopGrad}(P_{\text{src}}^{\text{welch}}, P_{\text{ref}})$
\STATE Scaling mask in amplitude domain: $M \leftarrow \sqrt{\dfrac{P_{\text{ref}}}{P_{\text{src}}^{\text{welch}}+\epsilon}}$
\STATE Align $M$ to FFT frequency bins: $\tilde{M}\leftarrow \textsc{InterpFreqBins}(M,\ \text{target}=\text{FreqBins}(L))$
\STATE Frequency-domain calibration:
$Y \leftarrow \Re\Big\{\text{IFFT}\big(\text{FFT}(X)\odot \tilde{M}\big)\Big\}$

\STATE \textbf{return} $Y$

\end{algorithmic}
\end{algorithm}

\clearpage
\section{Spectral Visualization of Structural Stratification in the Sleep Staging Task}
In the sleep staging task, we perform a spectral visualization analysis of the power spectral density (PSD) to investigate the structural stratification induced by different choices of $K$, with particular attention to the best-performing setting $K=8$. For each stratum, we visualize the normalized PSD curves of the samples assigned to that stratum, along with their corresponding variability ranges, in order to characterize within-stratum spectral variation.
When $K$ exceeds 2, clear and stable differences in spectral morphology already emerge across strata. These differences are primarily reflected in the distribution of low-frequency energy, the locations of dominant frequency bands, and the overall spectral decay patterns. In contrast, PSD curves within the same stratum remain highly consistent, indicating that samples grouped into the same stratum share similar spectral characteristics.
\begin{figure}[H]
    \centering
    \includegraphics[width=0.78\linewidth]{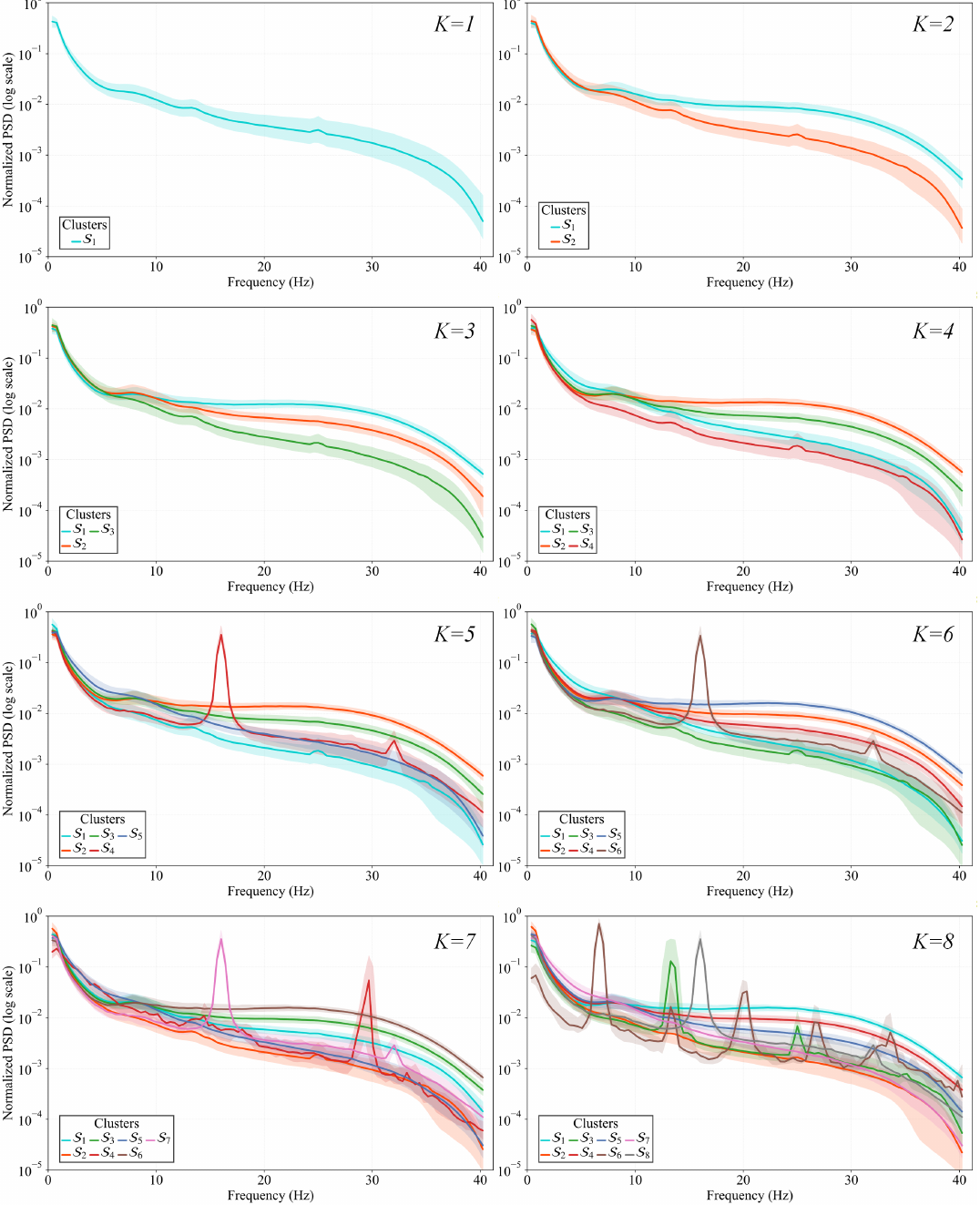}
    \caption{
 Spectral visualization of structural stratification in the sleep staging task under $K=8$.
    Each subfigure corresponds to one structural stratum and shows the normalized power spectral density (PSD) curves of the samples assigned to that stratum, together with their variability ranges.
    Different strata exhibit clearly distinct spectral profiles in terms of low-frequency energy concentration, dominant frequency bands, and overall spectral decay behavior, while samples within the same stratum demonstrate high spectral consistency.
    }
    \label{fig:clusters_overlay}
\end{figure}


\section{Anchor Matching Quality}
\subsection{Results on LODO Scenarios}
\begin{table}[H]
\centering
\setlength{\tabcolsep}{5.4pt}
\renewcommand{\arraystretch}{1.12}
\caption{
Impact of Anchor Matching Quality on Sleep Staging (LODO, Structural Granularity = 8).
The ``Rank'' represents the matching quality measured by the distance to the MAS template, where a lower rank indicates better quality.
I--V correspond to SHHS1, P2018, MROS1, MROS2, and MESA, respectively.
Each column leaves one domain out as the target domain (TD) and uses the remaining domains as source domains (SD).
}
\label{tab:rank_center_results}

\begin{tabular}{cc ccccc c}
\toprule
\multirow{2}{*}{\textbf{Rank}}
& \multirow{2}{*}{\textbf{Metric}}
& \textbf{II, III, IV, V}
& \textbf{I, III, IV, V}
& \textbf{I, II, IV, V}
& \textbf{I, II, III, V}
& \textbf{I, II, III, IV}
& \multirow{2}{*}{\textbf{AVG}} \\

&
& \textbf{I}
& \textbf{II}
& \textbf{III}
& \textbf{IV}
& \textbf{V}
& \\
\midrule

\multirow{2}{*}{\textbf{1}} & \textbf{ACC}
& $81.18 \pm 0.25$ & $75.68 \pm 0.67$ & $86.23 \pm 0.24$ & $87.70 \pm 0.37$ & $76.10 \pm 0.44$
& $\mathbf{81.38 \pm 0.39}$ \\
& \textbf{MF1}
& $71.81 \pm 0.27$ & $71.13 \pm 0.67$ & $71.16 \pm 0.32$ & $68.59 \pm 0.67$ & $63.16 \pm 0.50$
& $\mathbf{69.37 \pm 0.49}$ \\
\addlinespace[0.25em]

\multirow{2}{*}{\textbf{2}} & \textbf{ACC}
& $77.27 \pm 0.30$ & $70.40 \pm 0.82$ & $81.06 \pm 0.31$ & $85.33 \pm 0.43$ & $70.09 \pm 0.50$
& $\mathbf{76.83 \pm 0.47}$ \\
& \textbf{MF1}
& $67.75 \pm 0.28$ & $66.46 \pm 0.76$ & $65.42 \pm 0.32$ & $65.99 \pm 0.62$ & $57.29 \pm 0.48$
& $\mathbf{64.58 \pm 0.49}$ \\
\addlinespace[0.25em]

\multirow{2}{*}{\textbf{3}} & \textbf{ACC}
& $69.99 \pm 0.34$ & $66.83 \pm 0.86$ & $72.21 \pm 0.37$ & $80.58 \pm 0.57$ & $62.42 \pm 0.58$
& $\mathbf{70.41 \pm 0.54}$ \\
& \textbf{MF1}
& $61.84 \pm 0.31$ & $63.07 \pm 0.78$ & $55.65 \pm 0.34$ & $59.44 \pm 0.64$ & $48.51 \pm 0.48$
& $\mathbf{57.70 \pm 0.51}$ \\
\addlinespace[0.25em]

\multirow{2}{*}{\textbf{4}} & \textbf{ACC}
& $61.59 \pm 0.41$ & $61.70 \pm 0.97$ & $61.76 \pm 0.43$ & $75.31 \pm 0.66$ & $59.82 \pm 0.60$
& $\mathbf{64.04 \pm 0.61}$ \\
& \textbf{MF1}
& $54.38 \pm 0.36$ & $58.17 \pm 0.80$ & $45.83 \pm 0.37$ & $54.42 \pm 0.70$ & $45.61 \pm 0.48$
& $\mathbf{51.68 \pm 0.54}$ \\
\addlinespace[0.25em]

\multirow{2}{*}{\textbf{5}} & \textbf{ACC}
& $64.89 \pm 0.39$ & $60.63 \pm 1.00$ & $53.39 \pm 0.43$ & $69.89 \pm 0.69$ & $59.79 \pm 0.56$
& $\mathbf{61.72 \pm 0.61}$ \\
& \textbf{MF1}
& $53.86 \pm 0.37$ & $57.25 \pm 0.89$ & $35.45 \pm 0.44$ & $50.41 \pm 0.67$ & $44.85 \pm 0.49$
& $\mathbf{48.36 \pm 0.57}$ \\
\addlinespace[0.25em]

\multirow{2}{*}{\textbf{6}} & \textbf{ACC}
& $52.39 \pm 0.41$ & $53.77 \pm 1.04$ & $56.91 \pm 0.43$ & $72.68 \pm 0.67$ & $58.29 \pm 0.58$
& $\mathbf{58.81 \pm 0.63}$ \\
& \textbf{MF1}
& $45.82 \pm 0.38$ & $51.69 \pm 0.79$ & $40.54 \pm 0.38$ & $52.24 \pm 0.71$ & $43.57 \pm 0.47$
& $\mathbf{46.77 \pm 0.55}$ \\
\addlinespace[0.25em]

\multirow{2}{*}{\textbf{7}} & \textbf{ACC}
& $31.53 \pm 0.33$ & $36.52 \pm 0.96$ & $56.99 \pm 0.45$ & $67.88 \pm 0.72$ & $51.20 \pm 0.56$
& $\mathbf{48.82 \pm 0.60}$ \\
& \textbf{MF1}
& $16.84 \pm 0.27$ & $37.29 \pm 0.76$ & $39.27 \pm 0.42$ & $47.63 \pm 0.69$ & $35.09 \pm 0.46$
& $\mathbf{35.22 \pm 0.52}$ \\
\addlinespace[0.25em]

\multirow{2}{*}{\textbf{8}} & \textbf{ACC}
& $33.96 \pm 0.33$ & $40.32 \pm 0.98$ & $46.41 \pm 0.42$ & $61.85 \pm 0.76$ & $51.65 \pm 0.55$
& $\mathbf{46.84 \pm 0.61}$ \\
& \textbf{MF1}
& $22.09 \pm 0.33$ & $41.40 \pm 0.79$ & $19.67 \pm 0.34$ & $38.90 \pm 0.62$ & $37.29 \pm 0.42$
& $\mathbf{31.87 \pm 0.50}$ \\
\bottomrule
\end{tabular}
\label{tab:rank_lodo_g8}
\end{table}

\begin{table}[H]
\centering
\setlength{\tabcolsep}{5.4pt}
\renewcommand{\arraystretch}{1.12}
\caption{
Impact of Anchor Matching Quality on Sleep Staging (LODO, Structural Granularity = 5).
The ``Rank'' represents the matching quality measured by the distance to the MAS template, where a lower rank indicates better quality.
I--V correspond to SHHS1, P2018, MROS1, MROS2, and MESA, respectively.
Each column leaves one domain out as the target domain (TD) and uses the remaining domains as source domains (SD).
}
\label{tab:rank_center_results}

\begin{tabular}{cc ccccc c}
\toprule
\multirow{2}{*}{\textbf{Rank}}
& \multirow{2}{*}{\textbf{Metric}}
& \textbf{II, III, IV, V}
& \textbf{I, III, IV, V}
& \textbf{I, II, IV, V}
& \textbf{I, II, III, V}
& \textbf{I, II, III, IV}
& \multirow{2}{*}{\textbf{AVG}} \\

&
& \textbf{I}
& \textbf{II}
& \textbf{III}
& \textbf{IV}
& \textbf{V}
& \\
\midrule

\multirow{2}{*}{\textbf{1}} & \textbf{ACC}
& $79.68 \pm 0.27$ & $77.02 \pm 0.64$ & $86.46 \pm 0.22$ & $87.11 \pm 0.39$ & $75.98 \pm 0.46$
& $\mathbf{81.25 \pm 0.40}$ \\
& \textbf{MF1}
& $70.32 \pm 0.29$ & $72.99 \pm 0.60$ & $73.62 \pm 0.29$ & $68.71 \pm 0.63$ & $62.85 \pm 0.49$
& $\mathbf{69.70 \pm 0.46}$ \\
\addlinespace[0.25em]

\multirow{2}{*}{\textbf{2}} & \textbf{ACC}
& $79.41 \pm 0.28$ & $75.15 \pm 1.85$ & $80.92 \pm 0.31$ & $84.74 \pm 0.41$ & $69.79 \pm 0.51$
& $\mathbf{78.00 \pm 0.67}$ \\
& \textbf{MF1}
& $70.05 \pm 0.30$ & $66.24 \pm 2.01$ & $67.66 \pm 0.34$ & $62.80 \pm 0.53$ & $56.78 \pm 0.52$
& $\mathbf{64.71 \pm 0.74}$ \\
\addlinespace[0.25em]

\multirow{2}{*}{\textbf{3}} & \textbf{ACC}
& $78.60 \pm 0.28$ & $71.32 \pm 2.09$ & $70.93 \pm 0.40$ & $82.51 \pm 0.45$ & $64.66 \pm 0.57$
& $\mathbf{73.60 \pm 0.76}$ \\
& \textbf{MF1}
& $68.65 \pm 0.27$ & $61.27 \pm 2.12$ & $57.03 \pm 0.40$ & $59.79 \pm 0.52$ & $51.15 \pm 0.51$
& $\mathbf{59.58 \pm 0.76}$ \\
\addlinespace[0.25em]

\multirow{2}{*}{\textbf{4}} & \textbf{ACC}
& $79.72 \pm 0.26$ & $48.58 \pm 2.77$ & $67.94 \pm 0.44$ & $82.55 \pm 0.47$ & $65.12 \pm 0.60$
& $\mathbf{68.78 \pm 0.91}$ \\
& \textbf{MF1}
& $67.63 \pm 0.26$ & $43.42 \pm 2.13$ & $54.04 \pm 0.42$ & $60.76 \pm 0.56$ & $51.21 \pm 0.55$
& $\mathbf{55.41 \pm 0.78}$ \\
\addlinespace[0.25em]

\multirow{2}{*}{\textbf{5}} & \textbf{ACC}
& $77.46 \pm 0.29$ & $39.26 \pm 2.81$ & $53.88 \pm 0.42$ & $77.30 \pm 0.65$ & $57.95 \pm 0.62$
& $\mathbf{61.17 \pm 0.96}$ \\
& \textbf{MF1}
& $64.33 \pm 0.27$ & $33.55 \pm 1.97$ & $37.49 \pm 0.40$ & $56.94 \pm 0.67$ & $44.95 \pm 0.56$
& $\mathbf{47.45 \pm 0.77}$ \\
\bottomrule
\end{tabular}
\label{tab:rank_lodo_g5}
\end{table}

\begin{table}[H]
\centering
\setlength{\tabcolsep}{8pt} 
\renewcommand{\arraystretch}{1.12}
\caption{
Impact of Anchor Matching Quality on HAR (LODO, Structural Granularity = 8).
The ``Rank'' represents the matching quality measured by the distance to the MAS template, where a lower rank indicates better quality.
I--IV correspond to MotionSense, UCIHAR, USCHAD, and WISDM, respectively.
Each column leaves one domain out as the target domain (TD) and uses the remaining domains as source domains (SD).
}
\label{tab:rank_lodo_har_g8}

\begin{tabular}{cc cccc c}
\toprule
\multirow{2}{*}{\textbf{Rank}}
& \multirow{2}{*}{\textbf{Metric}}
& \textbf{II, III, IV}
& \textbf{I, III, IV}
& \textbf{I, II, IV}
& \textbf{I, II, III}
& \multirow{2}{*}{\textbf{AVG}} \\

&
& \textbf{I}
& \textbf{II}
& \textbf{III}
& \textbf{IV}
& \\
\midrule

\multirow{2}{*}{\textbf{1}} & \textbf{ACC}
& $98.48 \pm 0.09$ & $92.04 \pm 0.27$ & $86.54 \pm 0.23$ & $99.21 \pm 0.04$
& $\mathbf{94.07 \pm 0.16}$ \\
& \textbf{MF1}
& $98.59 \pm 0.08$ & $90.41 \pm 0.34$ & $89.14 \pm 0.20$ & $99.56 \pm 0.02$
& $\mathbf{94.43 \pm 0.16}$ \\
\addlinespace[0.25em]

\multirow{2}{*}{\textbf{2}} & \textbf{ACC}
& $98.73 \pm 0.05$ & $90.73 \pm 0.26$ & $85.04 \pm 0.15$ & $99.13 \pm 0.05$
& $\mathbf{93.41 \pm 0.13}$ \\
& \textbf{MF1}
& $98.82 \pm 0.05$ & $88.71 \pm 0.27$ & $87.82 \pm 0.16$ & $99.52 \pm 0.02$
& $\mathbf{93.72 \pm 0.13}$ \\
\addlinespace[0.25em]

\multirow{2}{*}{\textbf{3}} & \textbf{ACC}
& $99.12 \pm 0.08$ & $79.20 \pm 0.50$ & $80.96 \pm 0.25$ & $98.98 \pm 0.04$
& $\mathbf{89.57 \pm 0.22}$ \\
& \textbf{MF1}
& $99.18 \pm 0.07$ & $70.36 \pm 0.74$ & $84.01 \pm 0.23$ & $99.44 \pm 0.02$
& $\mathbf{88.25 \pm 0.27}$ \\
\addlinespace[0.25em]

\multirow{2}{*}{\textbf{4}} & \textbf{ACC}
& $99.34 \pm 0.05$ & $77.47 \pm 0.39$ & $77.55 \pm 0.34$ & $98.94 \pm 0.05$
& $\mathbf{88.33 \pm 0.21}$ \\
& \textbf{MF1}
& $99.39 \pm 0.05$ & $66.88 \pm 0.66$ & $80.50 \pm 0.33$ & $99.42 \pm 0.03$
& $\mathbf{86.55 \pm 0.27}$ \\
\addlinespace[0.25em]
\bottomrule
\end{tabular}
\end{table}

\subsection{Results on Independent Test Sets}
\begin{table}[H]
\centering
\caption{
Impact of Anchor Matching Quality on Sleep Staging (Independent Test Sets, Structural Granularity = 8).
Performance (mean $\pm$ std). The ``Rank'' indicates the distance to the MAS template (lower is better).
The last column (AVG) reports the average performance across all datasets.
}
\resizebox{\textwidth}{!}{
    \setlength{\tabcolsep}{3.5pt} 
    \renewcommand{\arraystretch}{1.2}
    
    \begin{tabular}{cc cccccc c}
    \toprule
    \textbf{Rank} & \textbf{Metric} & \textbf{ABC} & \textbf{CCSHS} & \textbf{CFS} & \textbf{HMC} & \textbf{ISRUC} & \textbf{Sleep-EDFx} & \textbf{AVG} \\
    \midrule

    \multirow{2}{*}{\textbf{1}} & \textbf{ACC}
    & $76.88 \pm 1.45$ & $88.43 \pm 0.50$ & $86.92 \pm 0.55$ & $77.48 \pm 1.87$ & $79.04 \pm 1.32$ & $81.73 \pm 1.25$
    & $\mathbf{81.75 \pm 1.16}$ \\
    
    & \textbf{MF1}
    & $71.96 \pm 1.24$ & $81.04 \pm 0.58$ & $77.12 \pm 0.65$ & $72.49 \pm 1.75$ & $74.76 \pm 1.16$ & $73.33 \pm 1.42$
    & $\mathbf{75.12 \pm 1.13}$ \\
    \addlinespace[0.3em]

    \multirow{2}{*}{\textbf{2}} & \textbf{ACC}
    & $70.46 \pm 1.72$ & $88.34 \pm 0.49$ & $86.95 \pm 0.56$ & $75.93 \pm 1.86$ & $77.89 \pm 1.40$ & $76.93 \pm 1.64$
    & $\mathbf{79.42 \pm 1.28}$ \\
    
    & \textbf{MF1}
    & $64.27 \pm 1.67$ & $79.60 \pm 0.60$ & $75.59 \pm 0.66$ & $69.52 \pm 1.71$ & $71.95 \pm 1.33$ & $67.25 \pm 1.77$
    & $\mathbf{71.36 \pm 1.29}$ \\
    \addlinespace[0.3em]

    \multirow{2}{*}{\textbf{3}} & \textbf{ACC}
    & $61.93 \pm 2.06$ & $83.43 \pm 0.66$ & $82.60 \pm 0.72$ & $69.70 \pm 1.92$ & $73.24 \pm 1.81$ & $70.12 \pm 2.11$
    & $\mathbf{73.50 \pm 1.55}$ \\
    
    & \textbf{MF1}
    & $55.85 \pm 1.72$ & $71.38 \pm 0.78$ & $68.89 \pm 0.75$ & $62.39 \pm 1.66$ & $66.02 \pm 1.55$ & $59.97 \pm 2.03$
    & $\mathbf{64.08 \pm 1.42}$ \\
    \addlinespace[0.3em]

    \multirow{2}{*}{\textbf{4}} & \textbf{ACC}
    & $53.87 \pm 2.38$ & $78.28 \pm 0.85$ & $77.29 \pm 0.90$ & $62.38 \pm 2.07$ & $68.82 \pm 2.10$ & $64.32 \pm 2.42$
    & $\mathbf{67.49 \pm 1.79}$ \\
    
    & \textbf{MF1}
    & $49.44 \pm 1.88$ & $64.41 \pm 0.87$ & $62.88 \pm 0.82$ & $55.29 \pm 1.61$ & $61.46 \pm 1.80$ & $54.82 \pm 2.09$
    & $\mathbf{58.05 \pm 1.51}$ \\
    \addlinespace[0.3em]

    \multirow{2}{*}{\textbf{5}} & \textbf{ACC}
    & $52.73 \pm 2.70$ & $81.51 \pm 0.88$ & $76.37 \pm 1.00$ & $61.01 \pm 2.53$ & $65.67 \pm 2.40$ & $67.06 \pm 2.53$
    & $\mathbf{67.39 \pm 2.01}$ \\
    
    & \textbf{MF1}
    & $50.17 \pm 2.00$ & $68.02 \pm 0.84$ & $63.13 \pm 0.85$ & $55.07 \pm 1.75$ & $59.60 \pm 1.89$ & $58.50 \pm 1.91$
    & $\mathbf{59.08 \pm 1.54}$ \\
    \addlinespace[0.3em]

    \multirow{2}{*}{\textbf{6}} & \textbf{ACC}
    & $48.51 \pm 2.36$ & $74.22 \pm 0.94$ & $72.05 \pm 1.04$ & $55.77 \pm 2.19$ & $64.01 \pm 2.37$ & $58.94 \pm 2.73$
    & $\mathbf{62.25 \pm 1.94}$ \\
    
    & \textbf{MF1}
    & $45.54 \pm 1.95$ & $60.23 \pm 0.85$ & $58.37 \pm 0.87$ & $49.99 \pm 1.59$ & $57.49 \pm 1.91$ & $50.78 \pm 2.23$
    & $\mathbf{53.73 \pm 1.57}$ \\
    \addlinespace[0.3em]

    \multirow{2}{*}{\textbf{7}} & \textbf{ACC}
    & $38.35 \pm 2.26$ & $62.53 \pm 1.07$ & $58.52 \pm 1.11$ & $40.70 \pm 2.18$ & $49.57 \pm 2.62$ & $47.70 \pm 2.62$
    & $\mathbf{49.56 \pm 1.98}$ \\
    
    & \textbf{MF1}
    & $36.37 \pm 2.05$ & $48.25 \pm 0.89$ & $44.88 \pm 0.97$ & $36.47 \pm 1.70$ & $45.02 \pm 2.17$ & $41.85 \pm 2.03$
    & $\mathbf{42.14 \pm 1.64}$ \\
    \addlinespace[0.3em]

    \multirow{2}{*}{\textbf{8}} & \textbf{ACC}
    & $33.83 \pm 2.16$ & $55.10 \pm 1.07$ & $51.88 \pm 1.01$ & $33.52 \pm 2.21$ & $41.22 \pm 2.71$ & $42.85 \pm 2.67$
    & $\mathbf{43.07 \pm 1.97}$ \\
    
    & \textbf{MF1}
    & $32.17 \pm 1.89$ & $43.41 \pm 0.96$ & $39.00 \pm 0.91$ & $30.57 \pm 1.72$ & $37.39 \pm 2.10$ & $38.11 \pm 1.65$
    & $\mathbf{36.78 \pm 1.54}$ \\

    \bottomrule
    \end{tabular}
}
\label{tab:rank_independent_granularity_8}
\end{table}

\begin{table}[H]
\centering
\caption{
Impact of Anchor Matching Quality on Sleep Staging (Independent Test Sets, Structural Granularity = 5).
Performance (mean $\pm$ std). The ``Rank'' indicates the distance to the MAS template (lower is better).
The last column (AVG) reports the average performance across all datasets.
}
\resizebox{\textwidth}{!}{
    \setlength{\tabcolsep}{4pt} 
    \renewcommand{\arraystretch}{1.2} 
    
    \begin{tabular}{cc cccccc c}
    \toprule
    \textbf{Rank} & \textbf{Metric} & \textbf{ABC} & \textbf{CCSHS} & \textbf{CFS} & \textbf{HMC} & \textbf{ISRUC} & \textbf{Sleep-EDFx} & \textbf{AVG} \\
    \midrule

    \multirow{2}{*}{\textbf{1}} & \textbf{ACC}
    & $77.35 \pm 1.48$ & $86.63 \pm 0.63$ & $85.63 \pm 0.62$ & $76.99 \pm 1.86$ & $78.31 \pm 1.35$ & $81.57 \pm 1.24$
    & $\mathbf{81.08 \pm 1.20}$ \\
    
    & \textbf{MF1}
    & $72.26 \pm 1.20$ & $79.01 \pm 0.69$ & $75.49 \pm 0.71$ & $71.95 \pm 1.74$ & $74.00 \pm 1.19$ & $73.10 \pm 1.39$
    & $\mathbf{74.30 \pm 1.15}$ \\
    \addlinespace[0.3em] 

    \multirow{2}{*}{\textbf{2}} & \textbf{ACC}
    & $71.14 \pm 1.64$ & $87.12 \pm 0.53$ & $86.07 \pm 0.59$ & $75.62 \pm 1.81$ & $77.99 \pm 1.27$ & $76.90 \pm 1.64$
    & $\mathbf{79.14 \pm 1.25}$ \\
    
    & \textbf{MF1}
    & $64.75 \pm 1.48$ & $77.67 \pm 0.62$ & $74.08 \pm 0.70$ & $69.41 \pm 1.57$ & $72.16 \pm 1.16$ & $66.88 \pm 1.79$
    & $\mathbf{70.83 \pm 1.22}$ \\
    \addlinespace[0.3em]

    \multirow{2}{*}{\textbf{3}} & \textbf{ACC}
    & $67.02 \pm 1.93$ & $85.42 \pm 0.53$ & $84.81 \pm 0.60$ & $72.42 \pm 1.82$ & $75.62 \pm 1.40$ & $73.67 \pm 1.85$
    & $\mathbf{76.49 \pm 1.36}$ \\
    
    & \textbf{MF1}
    & $59.88 \pm 1.61$ & $73.34 \pm 0.64$ & $70.62 \pm 0.67$ & $65.27 \pm 1.54$ & $68.46 \pm 1.26$ & $62.60 \pm 1.84$
    & $\mathbf{66.70 \pm 1.26}$ \\
    \addlinespace[0.3em]

    \multirow{2}{*}{\textbf{4}} & \textbf{ACC}
    & $62.47 \pm 2.39$ & $85.13 \pm 0.62$ & $82.32 \pm 0.77$ & $69.86 \pm 2.12$ & $72.06 \pm 1.99$ & $71.80 \pm 2.20$
    & $\mathbf{73.94 \pm 1.68}$ \\
    
    & \textbf{MF1}
    & $55.94 \pm 1.85$ & $70.72 \pm 0.70$ & $67.55 \pm 0.71$ & $61.54 \pm 1.62$ & $64.29 \pm 1.57$ & $60.91 \pm 1.91$
    & $\mathbf{63.49 \pm 1.39}$ \\
    \addlinespace[0.3em]

    \multirow{2}{*}{\textbf{5}} & \textbf{ACC}
    & $50.14 \pm 2.65$ & $76.65 \pm 0.98$ & $72.14 \pm 1.11$ & $56.76 \pm 2.45$ & $61.99 \pm 2.54$ & $58.65 \pm 3.04$
    & $\mathbf{62.72 \pm 2.13}$ \\
    
    & \textbf{MF1}
    & $46.90 \pm 2.09$ & $61.47 \pm 0.84$ & $58.35 \pm 0.94$ & $50.65 \pm 1.81$ & $55.75 \pm 1.95$ & $51.86 \pm 2.37$
    & $\mathbf{54.16 \pm 1.67}$ \\

    \bottomrule
    \end{tabular}
}
\label{tab:rank_independent_formatted}
\end{table}

\clearpage
\section{Sensitivity to Structural Granularity}
\subsection{LODO Evaluation}
\begin{table}[H]
\centering
\setlength{\tabcolsep}{4.2pt}
\renewcommand{\arraystretch}{1.12}
\caption{
Sensitivity to Structural Granularity on Sleep Staging (LODO).
Performance comparison with Structural Granularity varying from 1 to 8.
I--V correspond to SHHS1, P2018, MROS1, MROS2, and MESA, respectively.
Each column leaves one domain out as the target domain (TD) and uses the remaining domains as source domains (SD).
}
\label{tab:granularity_lodo_sleep}

\begin{tabular}{cc ccccc c}
\toprule
\multirow{2}{*}{\textbf{Granularity}}
& \multirow{2}{*}{\textbf{Metric}}
& \textbf{II, III, IV, V}
& \textbf{I, III, IV, V}
& \textbf{I, II, IV, V}
& \textbf{I, II, III, V}
& \textbf{I, II, III, IV}
& \multirow{2}{*}{\textbf{AVG}} \\

&
& \textbf{I}
& \textbf{II}
& \textbf{III}
& \textbf{IV}
& \textbf{V}
& \\
\midrule

\multirow{2}{*}{\textbf{1}} & \textbf{ACC}
& $79.76 \pm 0.26$ & $71.68 \pm 0.82$ & $85.36 \pm 0.23$ & $84.67 \pm 0.43$ & $80.36 \pm 0.38$
& $\mathbf{80.37 \pm 0.42}$ \\
& \textbf{MF1}
& $69.20 \pm 0.27$ & $63.64 \pm 0.74$ & $70.41 \pm 0.34$ & $62.65 \pm 0.59$ & $66.08 \pm 0.48$
& $\mathbf{66.40 \pm 0.48}$ \\
\addlinespace[0.25em]

\multirow{2}{*}{\textbf{2}} & \textbf{ACC}
& $82.72 \pm 0.23$ & $68.89 \pm 0.76$ & $86.29 \pm 0.22$ & $86.56 \pm 0.43$ & $78.72 \pm 0.40$
& $\mathbf{80.64 \pm 0.41}$ \\
& \textbf{MF1}
& $72.48 \pm 0.26$ & $58.47 \pm 0.79$ & $73.99 \pm 0.30$ & $66.60 \pm 0.73$ & $64.00 \pm 0.47$
& $\mathbf{67.11 \pm 0.51}$ \\
\addlinespace[0.25em]

\multirow{2}{*}{\textbf{3}} & \textbf{ACC}
& $80.77 \pm 0.24$ & $72.98 \pm 0.71$ & $81.92 \pm 0.33$ & $85.17 \pm 0.41$ & $79.95 \pm 0.44$
& $\mathbf{80.16 \pm 0.43}$ \\
& \textbf{MF1}
& $72.59 \pm 0.27$ & $65.57 \pm 0.64$ & $68.90 \pm 0.36$ & $69.39 \pm 0.63$ & $65.57 \pm 0.52$
& $\mathbf{68.40 \pm 0.48}$ \\
\addlinespace[0.25em]

\multirow{2}{*}{\textbf{4}} & \textbf{ACC}
& $83.07 \pm 0.22$ & $76.40 \pm 0.64$ & $85.47 \pm 0.24$ & $86.65 \pm 0.39$ & $77.74 \pm 0.46$
& $\mathbf{81.87 \pm 0.39}$ \\
& \textbf{MF1}
& $73.81 \pm 0.25$ & $72.24 \pm 0.61$ & $68.46 \pm 0.32$ & $64.21 \pm 0.62$ & $64.67 \pm 0.50$
& $\mathbf{68.68 \pm 0.46}$ \\
\addlinespace[0.25em]

\multirow{2}{*}{\textbf{5}} & \textbf{ACC}
& $79.68 \pm 0.27$ & $77.02 \pm 0.64$ & $86.46 \pm 0.22$ & $87.11 \pm 0.39$ & $75.98 \pm 0.46$
& $\mathbf{81.25 \pm 0.40}$ \\
& \textbf{MF1}
& $70.32 \pm 0.29$ & $72.99 \pm 0.60$ & $73.62 \pm 0.29$ & $68.71 \pm 0.63$ & $62.85 \pm 0.49$
& $\mathbf{69.70 \pm 0.46}$ \\
\addlinespace[0.25em]

\multirow{2}{*}{\textbf{6}} & \textbf{ACC}
& $81.73 \pm 0.24$ & $73.12 \pm 0.75$ & $85.18 \pm 0.22$ & $84.79 \pm 0.42$ & $79.02 \pm 0.41$
& $\mathbf{80.77 \pm 0.41}$ \\
& \textbf{MF1}
& $73.19 \pm 0.26$ & $66.32 \pm 0.72$ & $69.98 \pm 0.29$ & $64.66 \pm 0.65$ & $64.35 \pm 0.46$
& $\mathbf{67.70 \pm 0.48}$ \\
\addlinespace[0.25em]

\multirow{2}{*}{\textbf{7}} & \textbf{ACC}
& $80.84 \pm 0.26$ & $71.39 \pm 0.74$ & $86.27 \pm 0.24$ & $86.88 \pm 0.40$ & $79.94 \pm 0.40$
& $\mathbf{81.06 \pm 0.41}$ \\
& \textbf{MF1}
& $72.26 \pm 0.28$ & $63.71 \pm 0.59$ & $72.25 \pm 0.32$ & $68.49 \pm 0.60$ & $62.53 \pm 0.46$
& $\mathbf{67.85 \pm 0.45}$ \\
\addlinespace[0.25em]

\multirow{2}{*}{\textbf{8}} & \textbf{ACC}
& $81.18 \pm 0.25$ & $75.68 \pm 0.67$ & $86.23 \pm 0.24$ & $87.70 \pm 0.37$ & $76.10 \pm 0.44$
& $\mathbf{81.38 \pm 0.39}$ \\
& \textbf{MF1}
& $71.81 \pm 0.27$ & $71.13 \pm 0.67$ & $71.16 \pm 0.32$ & $69.59 \pm 0.67$ & $63.16 \pm 0.50$
& $\mathbf{69.37 \pm 0.49}$ \\
\bottomrule
\end{tabular}
\end{table}

\begin{table}[H]
\centering
\setlength{\tabcolsep}{8pt}
\renewcommand{\arraystretch}{1.12}
\caption{
Sensitivity to Structural Granularity on Arrhythmia Detection (LODO).
Performance comparison with Structural Granularity varying from 1 to 16.
I--IV correspond to PTB-XL, MITBIHADB, INCART, and CHAPMAN, respectively.
Each column leaves one domain out as the target domain (TD) and uses the remaining domains as source domains (SD).
}
\label{tab:granularity_lodo_arrhythmia}

\begin{tabular}{cc cccc c}
\toprule
\multirow{2}{*}{\textbf{Granularity}}
& \multirow{2}{*}{\textbf{Metric}}
& \textbf{II, III, IV}
& \textbf{I, III, IV}
& \textbf{I, II, IV}
& \textbf{I, II, III}
& \multirow{2}{*}{\textbf{AVG}} \\

&
& \textbf{I}
& \textbf{II}
& \textbf{III}
& \textbf{IV}
& \\
\midrule

\multirow{2}{*}{\textbf{1}} & \textbf{ACC}
& $61.62 \pm 0.18$ & $80.99 \pm 0.24$ & $80.92 \pm 0.22$ & $65.92 \pm 0.22$
& $\mathbf{72.36 \pm 0.22}$ \\
& \textbf{MF1}
& $69.56 \pm 0.17$ & $89.51 \pm 0.16$ & $89.28 \pm 0.14$ & $76.70 \pm 0.18$
& $\mathbf{81.26 \pm 0.16}$ \\
\addlinespace[0.25em]

\multirow{2}{*}{\textbf{2}} & \textbf{ACC}
& $54.84 \pm 0.19$ & $76.37 \pm 0.27$ & $81.58 \pm 0.21$ & $52.18 \pm 0.24$
& $\mathbf{66.24 \pm 0.23}$ \\
& \textbf{MF1}
& $58.94 \pm 0.21$ & $83.02 \pm 0.22$ & $88.78 \pm 0.14$ & $61.95 \pm 0.24$
& $\mathbf{73.17 \pm 0.20}$ \\
\addlinespace[0.25em]

\multirow{2}{*}{\textbf{4}} & \textbf{ACC}
& $55.16 \pm 0.18$ & $78.12 \pm 0.26$ & $77.37 \pm 0.23$ & $53.47 \pm 0.23$
& $\mathbf{66.03 \pm 0.23}$ \\
& \textbf{MF1}
& $58.26 \pm 0.21$ & $84.48 \pm 0.20$ & $85.82 \pm 0.16$ & $63.68 \pm 0.23$
& $\mathbf{73.06 \pm 0.20}$ \\
\addlinespace[0.25em]

\multirow{2}{*}{\textbf{8}} & \textbf{ACC}
& $59.12 \pm 0.19$ & $82.95 \pm 0.23$ & $82.94 \pm 0.21$ & $53.75 \pm 0.23$
& $\mathbf{69.69 \pm 0.22}$ \\
& \textbf{MF1}
& $64.76 \pm 0.20$ & $88.41 \pm 0.17$ & $89.56 \pm 0.14$ & $63.77 \pm 0.23$
& $\mathbf{76.63 \pm 0.19}$ \\
\addlinespace[0.25em]

\multirow{2}{*}{\textbf{16}} & \textbf{ACC}
& $55.24 \pm 0.19$ & $81.56 \pm 0.25$ & $81.05 \pm 0.21$ & $51.84 \pm 0.23$
& $\mathbf{67.42 \pm 0.22}$ \\
& \textbf{MF1}
& $59.89 \pm 0.21$ & $87.24 \pm 0.19$ & $88.43 \pm 0.14$ & $61.59 \pm 0.24$
& $\mathbf{74.29 \pm 0.20}$ \\
\bottomrule
\end{tabular}
\end{table}

\begin{table}[H]
\centering
\setlength{\tabcolsep}{7pt}
\renewcommand{\arraystretch}{1.12}
\caption{
Sensitivity to Structural Granularity on HAR (LODO).
Performance comparison with Structural Granularity varying from 1 to 8.
I--IV correspond to MotionSense, UCIHAR, USCHAD, and WISDM, respectively.
Each column leaves one domain out as the target domain (TD) and uses the remaining domains as source domains (SD).
}
\label{tab:granularity_lodo_har}

\begin{tabular}{cc cccc c}
\toprule
\multirow{2}{*}{\textbf{Granularity}}
& \multirow{2}{*}{\textbf{Metric}}
& \textbf{II, III, IV}
& \textbf{I, III, IV}
& \textbf{I, II, IV}
& \textbf{I, II, III}
& \multirow{2}{*}{\textbf{AVG}} \\

&
& \textbf{I}
& \textbf{II}
& \textbf{III}
& \textbf{IV}
& \\
\midrule

\multirow{2}{*}{\textbf{1}} & \textbf{ACC}
& $99.34 \pm 0.06$ & $83.91 \pm 0.36$ & $86.57 \pm 0.23$ & $98.15 \pm 0.07$
& $\mathbf{91.99 \pm 0.18}$ \\
& \textbf{MF1}
& $99.39 \pm 0.05$ & $78.48 \pm 0.52$ & $89.16 \pm 0.20$ & $98.99 \pm 0.04$
& $\mathbf{91.51 \pm 0.20}$ \\
\addlinespace[0.25em]

\multirow{2}{*}{\textbf{2}} & \textbf{ACC}
& $99.21 \pm 0.06$ & $88.68 \pm 0.32$ & $87.48 \pm 0.23$ & $98.28 \pm 0.06$
& $\mathbf{93.41 \pm 0.17}$ \\
& \textbf{MF1}
& $99.27 \pm 0.06$ & $85.75 \pm 0.42$ & $89.95 \pm 0.20$ & $99.05 \pm 0.04$
& $\mathbf{93.51 \pm 0.18}$ \\
\addlinespace[0.25em]

\multirow{2}{*}{\textbf{3}} & \textbf{ACC}
& $98.77 \pm 0.08$ & $89.82 \pm 0.30$ & $87.42 \pm 0.23$ & $98.06 \pm 0.07$
& $\mathbf{93.52 \pm 0.17}$ \\
& \textbf{MF1}
& $98.86 \pm 0.07$ & $87.40 \pm 0.39$ & $89.90 \pm 0.19$ & $98.94 \pm 0.04$
& $\mathbf{93.78 \pm 0.17}$ \\
\addlinespace[0.25em]

\multirow{2}{*}{\textbf{4}} & \textbf{ACC}
& $98.48 \pm 0.09$ & $92.04 \pm 0.27$ & $86.54 \pm 0.23$ & $99.21 \pm 0.04$
& $\mathbf{94.07 \pm 0.16}$ \\
& \textbf{MF1}
& $98.59 \pm 0.08$ & $90.41 \pm 0.34$ & $89.14 \pm 0.20$ & $99.56 \pm 0.02$
& $\mathbf{94.43 \pm 0.16}$ \\
\addlinespace[0.25em]

\multirow{2}{*}{\textbf{5}} & \textbf{ACC}
& $98.88 \pm 0.07$ & $89.81 \pm 0.31$ & $86.69 \pm 0.23$ & $98.44 \pm 0.06$
& $\mathbf{93.46 \pm 0.17}$ \\
& \textbf{MF1}
& $98.96 \pm 0.07$ & $87.37 \pm 0.40$ & $89.25 \pm 0.20$ & $99.14 \pm 0.03$
& $\mathbf{93.68 \pm 0.18}$ \\
\addlinespace[0.25em]

\multirow{2}{*}{\textbf{6}} & \textbf{ACC}
& $98.55 \pm 0.08$ & $88.96 \pm 0.31$ & $87.02 \pm 0.23$ & $98.12 \pm 0.07$
& $\mathbf{93.16 \pm 0.17}$ \\
& \textbf{MF1}
& $98.66 \pm 0.08$ & $86.44 \pm 0.40$ & $89.62 \pm 0.20$ & $98.97 \pm 0.04$
& $\mathbf{93.42 \pm 0.18}$ \\
\addlinespace[0.25em]

\multirow{2}{*}{\textbf{7}} & \textbf{ACC}
& $98.30 \pm 0.09$ & $91.89 \pm 0.27$ & $85.92 \pm 0.23$ & $98.57 \pm 0.06$
& $\mathbf{93.67 \pm 0.16}$ \\
& \textbf{MF1}
& $98.43 \pm 0.08$ & $90.25 \pm 0.34$ & $88.56 \pm 0.21$ & $99.22 \pm 0.03$
& $\mathbf{94.12 \pm 0.17}$ \\
\addlinespace[0.25em]

\multirow{2}{*}{\textbf{8}} & \textbf{ACC}
& $98.72 \pm 0.08$ & $88.82 \pm 0.31$ & $87.32 \pm 0.23$ & $98.53 \pm 0.06$
& $\mathbf{93.35 \pm 0.17}$ \\
& \textbf{MF1}
& $98.81 \pm 0.07$ & $86.06 \pm 0.41$ & $89.90 \pm 0.19$ & $99.19 \pm 0.03$
& $\mathbf{93.49 \pm 0.18}$ \\
\bottomrule
\end{tabular}
\end{table}

\subsection{Results on Independent Test Sets}
\begin{table}[H]
\centering
\caption{
Sensitivity to Structural Granularity on Sleep Staging (Independent Test Sets).
Performance comparison (mean $\pm$ std) with Structural Granularity varying from 1 to 8.
The last column (AVG) reports the average performance across all six independent datasets.
}
\label{tab:granularity_indep_sleep}

\resizebox{\textwidth}{!}{
    \setlength{\tabcolsep}{3.5pt} 
    \renewcommand{\arraystretch}{1.2} 
    
    \begin{tabular}{cc cccccc c}
    \toprule
    \textbf{Granularity} & \textbf{Metric} & \textbf{ABC} & \textbf{CCSHS} & \textbf{CFS} & \textbf{HMC} & \textbf{ISRUC} & \textbf{Sleep-EDFx} & \textbf{AVG} \\
    \midrule

    \multirow{2}{*}{\textbf{1}} & \textbf{ACC}
    & $75.17 \pm 1.58$ & $85.96 \pm 0.53$ & $85.82 \pm 0.53$ & $73.39 \pm 2.00$ & $74.79 \pm 1.55$ & $82.13 \pm 1.27$
    & $\mathbf{79.54 \pm 1.24}$ \\
    
    & \textbf{MF1}
    & $67.94 \pm 1.40$ & $76.65 \pm 0.68$ & $73.63 \pm 0.69$ & $66.94 \pm 1.90$ & $68.72 \pm 1.45$ & $72.81 \pm 1.44$
    & $\mathbf{71.12 \pm 1.26}$ \\
    \addlinespace[0.3em]

    \multirow{2}{*}{\textbf{2}} & \textbf{ACC}
    & $76.07 \pm 1.57$ & $85.77 \pm 0.55$ & $85.79 \pm 0.55$ & $72.18 \pm 1.97$ & $73.76 \pm 1.61$ & $82.70 \pm 1.19$
    & $\mathbf{79.38 \pm 1.24}$ \\
    
    & \textbf{MF1}
    & $68.91 \pm 1.41$ & $77.15 \pm 0.65$ & $74.50 \pm 0.69$ & $65.88 \pm 1.94$ & $67.66 \pm 1.58$ & $73.37 \pm 1.37$
    & $\mathbf{71.25 \pm 1.27}$ \\
    \addlinespace[0.3em]

    \multirow{2}{*}{\textbf{3}} & \textbf{ACC}
    & $76.67 \pm 1.54$ & $86.74 \pm 0.58$ & $86.07 \pm 0.59$ & $76.24 \pm 1.94$ & $77.95 \pm 1.37$ & $77.95 \pm 1.37$
    & $\mathbf{80.27 \pm 1.23}$ \\
    
    & \textbf{MF1}
    & $71.46 \pm 1.23$ & $78.88 \pm 0.65$ & $75.64 \pm 0.68$ & $71.61 \pm 1.74$ & $73.38 \pm 1.25$ & $73.38 \pm 1.25$
    & $\mathbf{74.06 \pm 1.13}$ \\
    \addlinespace[0.3em]

    \multirow{2}{*}{\textbf{4}} & \textbf{ACC}
    & $75.40 \pm 1.59$ & $87.50 \pm 0.50$ & $86.64 \pm 0.53$ & $75.75 \pm 1.93$ & $77.64 \pm 1.38$ & $81.59 \pm 1.24$
    & $\mathbf{80.75 \pm 1.20}$ \\
    
    & \textbf{MF1}
    & $69.19 \pm 1.37$ & $78.92 \pm 0.63$ & $75.23 \pm 0.67$ & $70.18 \pm 1.82$ & $72.48 \pm 1.30$ & $72.04 \pm 1.42$
    & $\mathbf{73.01 \pm 1.20}$ \\
    \addlinespace[0.3em]

    \multirow{2}{*}{\textbf{5}} & \textbf{ACC}
    & $77.35 \pm 1.48$ & $86.63 \pm 0.63$ & $85.63 \pm 0.62$ & $76.99 \pm 1.86$ & $78.31 \pm 1.35$ & $81.57 \pm 1.24$
    & $\mathbf{81.08 \pm 1.20}$ \\
    
    & \textbf{MF1}
    & $72.26 \pm 1.20$ & $79.01 \pm 0.69$ & $75.49 \pm 0.71$ & $71.95 \pm 1.74$ & $74.00 \pm 1.19$ & $73.10 \pm 1.39$
    & $\mathbf{74.30 \pm 1.15}$ \\
    \addlinespace[0.3em]

    \multirow{2}{*}{\textbf{6}} & \textbf{ACC}
    & $75.04 \pm 1.62$ & $86.33 \pm 0.53$ & $86.33 \pm 0.51$ & $73.82 \pm 1.96$ & $74.72 \pm 1.59$ & $81.74 \pm 1.33$
    & $\mathbf{79.66 \pm 1.26}$ \\
    
    & \textbf{MF1}
    & $67.94 \pm 1.35$ & $77.52 \pm 0.66$ & $74.74 \pm 0.69$ & $67.77 \pm 1.88$ & $68.34 \pm 1.48$ & $72.79 \pm 1.53$
    & $\mathbf{71.52 \pm 1.27}$ \\
    \addlinespace[0.3em]

    \multirow{2}{*}{\textbf{7}} & \textbf{ACC}
    & $76.58 \pm 1.52$ & $88.11 \pm 0.51$ & $87.05 \pm 0.54$ & $77.15 \pm 1.89$ & $78.37 \pm 1.35$ & $83.91 \pm 1.05$
    & $\mathbf{81.86 \pm 1.14}$ \\
    
    & \textbf{MF1}
    & $70.70 \pm 1.26$ & $80.24 \pm 0.60$ & $76.32 \pm 0.67$ & $71.60 \pm 1.72$ & $73.03 \pm 1.21$ & $76.69 \pm 1.24$
    & $\mathbf{74.76 \pm 1.12}$ \\
    \addlinespace[0.3em]

    \multirow{2}{*}{\textbf{8}} & \textbf{ACC}
    & $76.88 \pm 1.45$ & $88.43 \pm 0.50$ & $86.92 \pm 0.55$ & $77.48 \pm 1.87$ & $79.04 \pm 1.32$ & $81.73 \pm 1.25$
    & $\mathbf{81.75 \pm 1.16}$ \\
    
    & \textbf{MF1}
    & $71.96 \pm 1.24$ & $81.04 \pm 0.58$ & $77.12 \pm 0.65$ & $72.49 \pm 1.75$ & $74.76 \pm 1.16$ & $73.33 \pm 1.42$
    & $\mathbf{75.12 \pm 1.13}$ \\

    \bottomrule
    \end{tabular}
}
\end{table}

\clearpage
\section{Additional Results with Accuracy}
\begin{table}[H] 
\centering
\caption{
Sleep staging performance (Accuracy) under leave-one-domain-out (LODO) evaluation and external target domain testing.
LODO explains average performance across five source-domain splits.
External results are evaluated on six unseen target datasets.
All values are reported as mean $\pm$ std over five runs.
}
\label{tab:sleep_results_acc_fixed}

\resizebox{\textwidth}{!}{
    \setlength{\tabcolsep}{4.0pt}
    \begin{tabular}{lcccccccc}
    \toprule
    \multirow{2}{*}{\textbf{Method}}
    & \multicolumn{1}{c}{\textbf{LODO}}
    & \multicolumn{7}{c}{\textbf{External Target Domains}} \\
    \cmidrule(lr){2-2} \cmidrule(lr){3-9}
    & \textbf{Avg}
    & \textbf{ABC}
    & \textbf{CCSHS}
    & \textbf{CFS}
    & \textbf{HMC}
    & \textbf{ISRUC}
    & \textbf{Sleep-EDFx}
    & \textbf{Avg} \\
    \midrule

    Baseline
    & 73.49 $\pm$ 0.57
    & 76.57 $\pm$ 1.63
    & 84.83 $\pm$ 0.70
    & 84.72 $\pm$ 0.62
    & 71.61 $\pm$ 2.12
    & 74.57 $\pm$ 1.53
    & 80.52 $\pm$ 1.35
    & 78.80 $\pm$ 1.33 \\

    LG-SleepNet
    & 67.80 $\pm$ 0.03
    & 73.32 $\pm$ 0.12
    & 79.44 $\pm$ 0.05
    & 81.00 $\pm$ 0.04
    & 70.92 $\pm$ 0.13
    & 73.89 $\pm$ 0.13
    & 62.21 $\pm$ 0.10
    & 73.46 $\pm$ 0.10 \\

    SleepEEGNet
    & 59.59 $\pm$ 0.03
    & 61.49 $\pm$ 0.13
    & 63.33 $\pm$ 0.06
    & 67.45 $\pm$ 0.05
    & 62.86 $\pm$ 0.13
    & 68.46 $\pm$ 0.14
    & 58.41 $\pm$ 0.10
    & 63.67 $\pm$ 0.10 \\

    IRM
    & 70.94 $\pm$ 0.60
    & 73.90 $\pm$ 1.80
    & 81.73 $\pm$ 0.75
    & 82.91 $\pm$ 0.64
    & 70.32 $\pm$ 2.17
    & 73.81 $\pm$ 1.67
    & 76.81 $\pm$ 1.53
    & 76.58 $\pm$ 1.43 \\

    MMD
    & 74.97 $\pm$ 0.47
    & 76.81 $\pm$ 1.64
    & 85.28 $\pm$ 0.67
    & 85.16 $\pm$ 0.60
    & 72.77 $\pm$ 2.10
    & 75.37 $\pm$ 1.49
    & 80.16 $\pm$ 1.37
    & 79.26 $\pm$ 1.31 \\

    CORAL
    & 72.91 $\pm$ 0.52
    & 76.55 $\pm$ 1.63
    & 85.96 $\pm$ 0.64
    & 85.66 $\pm$ 0.59
    & 73.06 $\pm$ 2.12
    & 75.46 $\pm$ 1.46
    & 79.87 $\pm$ 1.39
    & 79.43 $\pm$ 1.31 \\

    SleepDG
    & 74.88 $\pm$ 0.50
    & 76.82 $\pm$ 1.62
    & 86.61 $\pm$ 0.60
    & 86.16 $\pm$ 0.56
    & 72.61 $\pm$ 2.15
    & 75.19 $\pm$ 1.47
    & 80.53 $\pm$ 1.41
    & 79.65 $\pm$ 1.30 \\

    \midrule
    \textbf{Ours}
    & \textbf{81.38 $\pm$ 0.39}
    & \textbf{76.88 $\pm$ 1.45}
    & \textbf{88.43 $\pm$ 0.50}
    & \textbf{86.92 $\pm$ 0.55}
    & \textbf{77.48 $\pm$ 1.87}
    & \textbf{79.04 $\pm$ 1.32}
    & \textbf{81.73 $\pm$ 1.25}
    & \textbf{81.75 $\pm$ 1.16} \\
    \bottomrule
    \end{tabular}
}
\end{table}
\begin{table}[H]
\centering
\caption{Performance comparison (Accuracy $\pm$ Std). 
The table is organized into two panels: Arrhythmia (top) and HAR (bottom). 
SD: Source Domains, TD: Target Domain. 
The best results are highlighted in bold.}
\label{tab:ecg_har_acc_results}

\setlength{\tabcolsep}{9pt} 
\renewcommand{\arraystretch}{1.2} 

\begin{tabular}{l ccccc}
\toprule
\multicolumn{6}{c}{\textbf{Arrhythmia}} \\
\cmidrule(lr){1-6}

\textbf{SD} & \small II, III, IV & \small I, III, IV & \small I, II, IV & \small I, II, III &  \\
\textbf{TD} & \textbf{I} & \textbf{II} & \textbf{III} & \textbf{IV} & \textbf{Avg} \\
\cmidrule(lr){1-1} \cmidrule(lr){2-5} \cmidrule(lr){6-6}

Baseline & $56.77 \pm 0.18$ & $83.06 \pm 0.23$ & $74.71 \pm 0.23$ & $49.95 \pm 0.24$ & $66.12 \pm 0.22$ \\
CORAL    & $55.55 \pm 0.18$ & $\mathbf{83.26 \pm 0.23}$ & $78.62 \pm 0.22$ & $48.65 \pm 0.23$ & $66.52 \pm 0.22$ \\
MMD      & $53.85 \pm 0.18$ & $63.17 \pm 0.30$ & $49.86 \pm 0.27$ & $59.80 \pm 0.23$ & $56.67 \pm 0.25$ \\
IRM      & $54.62 \pm 0.19$ & $82.77 \pm 0.23$ & $\mathbf{81.59 \pm 0.21}$ & $46.66 \pm 0.23$ & $66.41 \pm 0.22$ \\
\textbf{Ours} & $\mathbf{61.62 \pm 0.18}$ & $80.99 \pm 0.24$ & $80.92 \pm 0.22$ & $\mathbf{65.92 \pm 0.22}$ & $\mathbf{72.36 \pm 0.22}$ \\

\midrule[\heavyrulewidth] 
\addlinespace[0.5em]      

\multicolumn{6}{c}{\textbf{Human Activity Recognition}} \\
\cmidrule(lr){1-6}

\textbf{SD} & \small II, III, IV & \small I, III, IV & \small I, II, IV & \small I, II, III &  \\
\textbf{TD} & \textbf{I} & \textbf{II} & \textbf{III} & \textbf{IV} & \textbf{Avg} \\
\cmidrule(lr){1-1} \cmidrule(lr){2-5} \cmidrule(lr){6-6}

Baseline & $84.84 \pm 0.25$ & $86.59 \pm 0.34$ & $65.86 \pm 0.31$ & $96.49 \pm 0.09$ & $83.45 \pm 0.25$ \\
CORAL    & $84.99 \pm 0.25$ & $94.10 \pm 0.23$ & $78.36 \pm 0.27$ & $96.96 \pm 0.08$ & $88.60 \pm 0.21$ \\
MMD      & $75.34 \pm 0.29$ & $94.25 \pm 0.23$ & $71.19 \pm 0.31$ & $94.98 \pm 0.11$ & $83.94 \pm 0.24$ \\
IRM      & $92.16 \pm 0.18$ & $\mathbf{94.38 \pm 0.23}$ & $80.77 \pm 0.26$ & $95.82 \pm 0.10$ & $90.78 \pm 0.19$ \\
\textbf{Ours} & $\mathbf{98.48 \pm 0.09}$ & $92.04 \pm 0.27$ & $\mathbf{86.54 \pm 0.23}$ & $\mathbf{99.21 \pm 0.04}$ & $\mathbf{94.07 \pm 0.16}$ \\

\bottomrule
\end{tabular}
\end{table}
\clearpage
\section{Supplementary LODO Results for Sleep Staging}
\begin{table}[H]
\centering
\caption{
Performance Comparison of Different Methods on Sleep Staging (LODO).
I--V correspond to SHHS1, P2018, MROS1, MROS2, and MESA, respectively.
Each column leaves one domain out as the target domain (TD) and uses the remaining domains as source domains (SD).
ACC and Macro-F1 (MF1) are reported as mean $\pm$ std.
}
\label{tab:method_sleepstage_lodo_comparison}

\resizebox{\textwidth}{!}{
    \setlength{\tabcolsep}{4.5pt} 
    \renewcommand{\arraystretch}{1.15} 
    
    \begin{tabular}{cc ccccc c}
    \toprule
    \multirow{2}{*}{\textbf{Method}}
    & \multirow{2}{*}{\textbf{Metric}}
    & \textbf{II, III, IV, V}
    & \textbf{I, III, IV, V}
    & \textbf{I, II, IV, V}
    & \textbf{I, II, III, V}
    & \textbf{I, II, III, IV}
    & \multirow{2}{*}{\textbf{AVG}} \\
    
    &
    & \textbf{I}
    & \textbf{II}
    & \textbf{III}
    & \textbf{IV}
    & \textbf{V}
    & \\
    \midrule
    
    \multirow{2}{*}{Baseline} & \textbf{ACC}
    & $76.58 \pm 0.31$ & $69.34 \pm 0.76$ & $85.04 \pm 0.25$ & $82.21 \pm 0.73$ & $54.26 \pm 0.80$
    & $73.49 \pm 0.57$ \\
    & \textbf{MF1}
    & $66.40 \pm 0.33$ & $59.44 \pm 0.74$ & $69.09 \pm 0.36$ & $65.90 \pm 0.79$ & $37.98 \pm 1.00$
    & $59.76 \pm 0.64$ \\
    \addlinespace[0.3em]
    
    \multirow{2}{*}{LG-SleepNet} & \textbf{ACC}
    & $71.79 \pm 0.02$ & $43.21 \pm 0.05$ & $79.56 \pm 0.02$ & $82.02 \pm 0.03$ & $62.41 \pm 0.03$
    & $67.80 \pm 0.03$ \\
    & \textbf{MF1}
    & $61.48 \pm 0.02$ & $41.91 \pm 0.05$ & $68.48 \pm 0.03$ & $65.04 \pm 0.06$ & $50.34 \pm 0.04$
    & $57.45 \pm 0.04$ \\
    \addlinespace[0.3em]
    
    \multirow{2}{*}{SleepEEGNet} & \textbf{ACC}
    & $63.68 \pm 0.02$ & $60.63 \pm 0.05$ & $71.42 \pm 0.02$ & $67.30 \pm 0.04$ & $34.90 \pm 0.03$
    & $59.59 \pm 0.03$ \\
    & \textbf{MF1}
    & $61.55 \pm 0.02$ & $60.95 \pm 0.05$ & $61.02 \pm 0.03$ & $53.24 \pm 0.05$ & $29.34 \pm 0.03$
    & $53.22 \pm 0.04$ \\
    \addlinespace[0.3em]
    
    \multirow{2}{*}{IRM} & \textbf{ACC}
    & $76.18 \pm 0.27$ & $68.32 \pm 0.79$ & $83.54 \pm 0.27$ & $75.01 \pm 0.91$ & $51.67 \pm 0.76$
    & $70.94 \pm 0.60$ \\
    & \textbf{MF1}
    & $62.67 \pm 0.32$ & $58.26 \pm 0.75$ & $67.13 \pm 0.33$ & $60.22 \pm 0.79$ & $34.00 \pm 0.97$
    & $56.46 \pm 0.63$ \\
    \addlinespace[0.3em]
    
    \multirow{2}{*}{MMD} & \textbf{ACC}
    & $76.87 \pm 0.29$ & $77.58 \pm 0.57$ & $85.25 \pm 0.23$ & $85.95 \pm 0.46$ & $49.19 \pm 0.80$
    & $74.97 \pm 0.47$ \\
    & \textbf{MF1}
    & $65.89 \pm 0.32$ & $75.29 \pm 0.56$ & $71.16 \pm 0.34$ & $68.44 \pm 0.68$ & $33.09 \pm 1.02$
    & $62.77 \pm 0.58$ \\
    \addlinespace[0.3em]
    
    \multirow{2}{*}{CORAL} & \textbf{ACC}
    & $74.96 \pm 0.33$ & $71.49 \pm 0.75$ & $84.43 \pm 0.26$ & $85.73 \pm 0.45$ & $47.96 \pm 0.81$
    & $72.91 \pm 0.52$ \\
    & \textbf{MF1}
    & $64.04 \pm 0.33$ & $63.94 \pm 0.68$ & $70.96 \pm 0.34$ & $67.15 \pm 0.64$ & $31.81 \pm 1.06$
    & $59.58 \pm 0.61$ \\
    \addlinespace[0.3em]
    
    \multirow{2}{*}{SleepDG} & \textbf{ACC}
    & $77.88 \pm 0.30$ & $70.53 \pm 0.75$ & $85.69 \pm 0.24$ & $86.09 \pm 0.42$ & $54.22 \pm 0.77$
    & $74.88 \pm 0.50$ \\
    & \textbf{MF1}
    & $65.72 \pm 0.31$ & $62.24 \pm 0.68$ & $70.82 \pm 0.35$ & $68.65 \pm 0.63$ & $37.19 \pm 1.00$
    & $60.92 \pm 0.59$ \\
    \addlinespace[0.3em]
    
    \multirow{2}{*}{\textbf{Ours}} & \textbf{ACC}
    & $\mathbf{81.18 \pm 0.25}$ & $\mathbf{75.68 \pm 0.67}$ & $\mathbf{86.23 \pm 0.24}$ & $\mathbf{87.70 \pm 0.37}$ & $\mathbf{76.10 \pm 0.44}$
    & $\mathbf{81.38 \pm 0.39}$ \\
    & \textbf{MF1}
    & $\mathbf{71.81 \pm 0.27}$ & $\mathbf{71.13 \pm 0.67}$ & $\mathbf{71.16 \pm 0.32}$ & $\mathbf{69.59 \pm 0.67}$ & $\mathbf{63.16 \pm 0.50}$
    & $\mathbf{69.37 \pm 0.49}$ \\
    
    \bottomrule
    \end{tabular}
}
\end{table}

\end{document}